\documentclass[english]{IEEEtran}
\usepackage[T1]{fontenc}
\usepackage[latin9]{inputenc}
\usepackage{array}
\usepackage{refstyle}
\usepackage{units}
\usepackage{url}
\usepackage{graphicx}

\makeatletter

\AtBeginDocument{\providecommand\figref[1]{\ref{fig:#1}}}
\AtBeginDocument{\providecommand\tabref[1]{\ref{tab:#1}}}
\providecommand{\tabularnewline}{\\}
\RS@ifundefined{subsecref}
  {\newref{subsec}{name = \RSsectxt}}
  {}
\RS@ifundefined{thmref}
  {\def\RSthmtxt{theorem~}\newref{thm}{name = \RSthmtxt}}
  {}
\RS@ifundefined{lemref}
  {\def\RSlemtxt{lemma~}\newref{lem}{name = \RSlemtxt}}
  {}

\usepackage{url}
\usepackage{hyperref}

\renewcommand{\tabref}{\Tabref}
\renewcommand{\figref}{\Figref}
\hypersetup {
    colorlinks,
    linkcolor={red},
    citecolor={black},
    urlcolor={blue}
}

\@ifundefined{showcaptionsetup}{}{%
 \PassOptionsToPackage{caption=false}{subfig}}
\usepackage{subfig}
\makeatother

\usepackage{babel}
\begin{document}

\title{Joint Attention in Autonomous Driving (JAAD)}

\author{Iuliia Kotseruba, Amir Rasouli and John K. Tsotsos\\
\{yulia\_k, aras, tsotsos\}@cse.yorku.ca}
\maketitle
\begin{abstract}
In this paper we present a novel dataset for a critical aspect of
autonomous driving, the joint attention that must occur between drivers
and of pedestrians, cyclists or other drivers. This dataset is produced
with the intention of demonstrating the behavioral variability of
traffic participants. We also show how visual complexity of the behaviors
and scene understanding is affected by various factors such as different
weather conditions, geographical locations, traffic and demographics
of the people involved. The ground truth data conveys information
regarding the location of participants (bounding boxes), the physical
conditions (e.g. lighting and speed) and the behavior of the parties
involved.
\end{abstract}

\section{Introduction}

Autonomous driving has been a topic of interest for decades. Implementing
autonomous vehicles can have a great economic and social impacts including
reducing the cost of driving, increasing fuel efficiency and safety,
enabling transportation for non-drivers and reducing the stress of
driving by allowing motorists to rest and work while traveling \cite{Litm2015}.
As for the macroeconomic impacts, it is estimated that autonomous
vehicles industry and related software and hardware technologies will
account for a market size of more than 40 billion dollars by 2030 \cite{Rol2014}.

Partial autonomy has long been used in commercial vehicles in the
form of technologies such as cruise control, park assist, automatic
braking, etc. Fully autonomous vehicles also have been successfully
developed and tested under certain conditions. For example, the DARPA
challenge 2005 set the task of autonomously driving a 7.32 miles predefined
terrain in the deserts of Nevada. Out of 23 final contestants 4 cars
successfully completed the course within the allowable time limit
(10 hours) while driving fully autonomously \cite{SebastianThrun2006}.

Despite such success stories in autonomous control systems, designing
fully autonomous vehicles for urban environments still remains an
unsolved problem. Aside from challenges associated with developing
suitable infrastructures and regulating the autonomous behaviors \cite{Litm2015},
in order to be usable in urban environments autonomous cars must have
a high level of precision and meet very high safety standards \cite{Kalra2006}.

Today one of the major dilemmas faced by autonomous vehicles is how
to interact with the environment including infrastructure, cars, drivers
or pedestrians \cite{Knight2015,Gomes2014,Silberg2012}. The lapses
in communication can be a source of numerous erroneous behaviors \cite{Anthony2016}
such as failure to predict the movement of other vehicles \cite{Richtel2016,Google2016} or to respond to unexpected behaviors of other drivers \cite{Knight2013}.

The impact of perceptual failures on the behavior of an autonomous
car is also evident in the 2015 annual report on Google's self-driving
car \cite{Google2015}. This report is based on testing self-driving
cars for more than $424,000$ miles of driving on public roads including
both highways and streets. Throughout these trials, a total of 341
disengagements occurred in which the driver had to take over the car,
and about $90\%$ of the cases occurred in busy streets. The interesting
implication here is that over $\nicefrac{1}{3}$ of the disengagements
were due to``perception discrepancy'' in which the vehicle was unable
to understand its environment and about 10\% of the cases were due
to incorrect prediction of traffic participants and inability to respond
to reckless behaviors.

There have been a number of recent developments to address these issues.
A natural solution is establishing wireless communication between
traffic participants. This approach has been tested for a number of
years using cellular technology \cite{Silberg2012,V2V2014a}. This
technique enables vehicle to vehicle (V2V) and vehicle to infrastructure
(V2I) communication allowing tasks such as Cooperative Adaptive Cruise
Control (CACC), improving positioning technologies such as GPS, and
intelligent speed adoption in various roads. Peer to peer traffic
communication is expected to enter the market by 2019.

Although V2V and V2I communications are deemed to solve a number of
issues in autonomous driving, they also have a number of drawbacks.
This technology relies heavily on cellular technology which is costly
and has much lower reliability compared to traditional sensors such
as radars and cameras. In addition, communication highly depends on
all parties functioning properly. A malfunction in any communication
device in any of the systems involved can lead to catastrophic safety
issues.

Maintaining communication with pedestrians is even more important
for safe autonomous driving. Inattention from both drivers and pedestrians
regardless of their knowledge of the vehicle code is one of the major
reasons for traffic accidents, most of which involve pedestrians at
crosswalk locations \cite{Ragland2007}.

Honda recently released a new technology similar to V2V communication
that attempts to establish a connection with pedestrians through their
cellular phones \cite{Honda2016a}. Using this method, the pedestrian's
phone broadcasts its position warning the autonomous car that a pedestrian
is about to cross the street so the car can respond accordingly. This
technology also can go one step further and inform the car about the
state of the pedestrian, for instance, whether he/she is listening
to music, texting or is on a call. Given the technological and regulatory
obstacles to developing such technologies, using them in the near
future does not seem feasible.

In late 2015 Google patented a different technology to communicate
with pedestrians using a visual interface called pedestrian notifications  \cite{urmson2015pedestrian}. 
In this approach, the Google car estimates the trajectories of pedestrian
movements. If the car finds the behavior of a pedestrian to be uncertain
(i.e. cannot decide whether or not he/she is crossing the street),
it notifies the corresponding pedestrian about the action it is about
to take using a screen installed on the front hood of the car. Another
proposed option for communication is via a sound device or other kinds
of physical devices (possibly a robotic arm). This technology has
been criticized for being distracting and lacking the ability to efficiently communicate if more than one pedestrian is involved.

Given the problems associated with establishing explicit communication
with other vehicles and pedestrians, Nissan, in their latest development,
announced a passive method of dealing with uncertainties in the behavior
of pedestrians \cite{Stern2015,Jones2016}, in attempt to understand
human drivers and pedestrians behaviors in various traffic scenarios.
The main objective of this work is to passively predict pedestrian
behavior using visual input and only use an interface, e.g. a green
light, to inform them about the intention of the autonomous car.

Toyota in partnership with MIT and Stanford recently announced using
a similar passive approach toward autonomous driving \cite{Ackerman2015}.
Information such as the type of equipment the pedestrian carries,
his/her pose, direction of motion and behavior as well as the human
driver's reactions to events are extracted from videos of traffic
situations and their 3D reconstructions. This information is used
to design a control system for determining what course of action to
take in given situations. At present no data resulting from this study
has been released and very little information is available about the
type of autonomous behavior the scientists are seeking to obtain.

Given the importance of pedestrian safety, multiple studies were conducted
in the last several decades to find factors that influence decisions
and behavior of traffic participants. For example, a driver's pedestrian
awareness can be measured based on whether the driver is decelerating
upon seeing a pedestrian (\cite{Akamatsu2003}, \cite{Fukagawa2013}, \cite{Phan2014}).
Several recent studies also point out that pedestrian's behavior,
such as establishing eye-contact, smiling and other forms of non-verbal
communication, can have a significant impact on the driver's actions
(\cite{Ren1026}, \cite{Gueguen2016}). Although a majority of these
studies are aimed at developing better infrastructure and traffic
regulations, their conclusions are relevant for autonomous driving
as well.

In an attempt to better understand the problem of vehicle to vehicle
(V2V) and vehicle to pedestrian (V2P) communication in the autonomous
driving context we suggest viewing it as an instance of joint attention
and discuss why existing approaches may not be adequate in this context.
We propose a novel dataset that highlights the visual and behavioral
complexity of traffic scene understanding and is potentially valuable
for studying the joint attention issues.

\section{Autonomous driving and joint attention}

According to a common definition, joint attention is the ability to
detect and influence an observable attentional behavior of another
agent in social interaction and acknowledge them as an intentional
agent \cite{Kaplan2006a}. However, it is important to note that joint
attention is more than simultaneous looking, attention detection and
social coordination, but also includes an intentional understanding
of the observed behavior of others.

Since joint attention is a prerequisite for efficient communication,
it has been gaining increasing interest in the fields of robotics
and human-robot interaction. Kismet \cite{Breazeal1999b} and Cog \cite{Scassellati1999},
both built at MIT in the late 1990s, were some of the first successes
in social robotics. These robots were able to maintain and follow
eye gaze, reacted to the behavior of their caregivers and recognized
simple gestures such as declarative pointing. More recent work in
this area is likewise concerned with gaze following \cite{Shon2005,Fasel2002,Scassellati1999},
pointing \cite{Hafner2005,Scassellati1999} and reaching \cite{Doniec2006},
turn-taking \cite{Andry2001} and social referencing \cite{Boucenna2011}.
With a few exceptions \cite{May2015,Ishiguro2002}, almost all joint
attention scenarios are implemented with stationary robots or robotic
heads according to a recent comprehensive survey \cite{Ferreira2014}.

\begin{figure*}
\noindent \begin{centering}
\subfloat[\label{fig:woman_crossing}]{\includegraphics[scale=0.25]{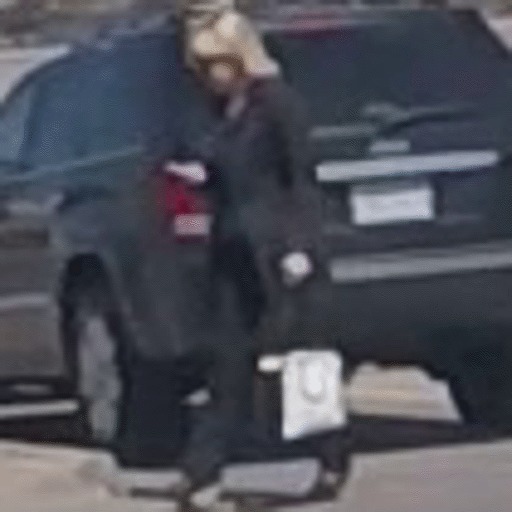}\includegraphics[scale=0.25]{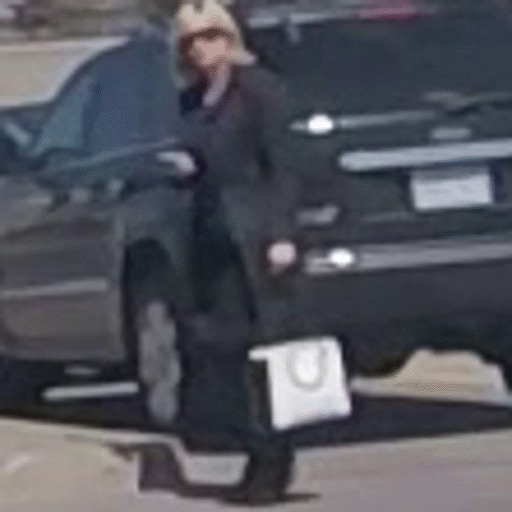}\includegraphics[scale=0.25]{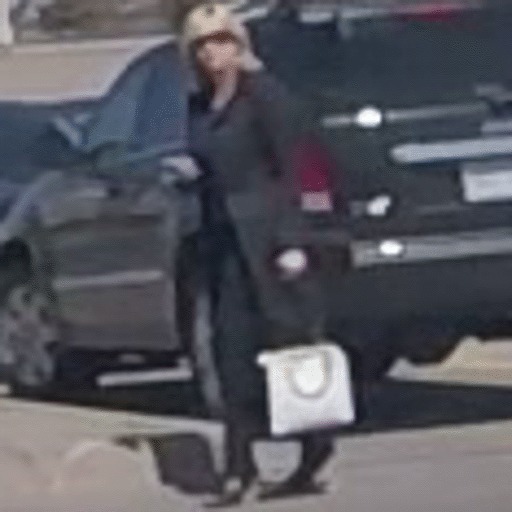}\includegraphics[scale=0.25]{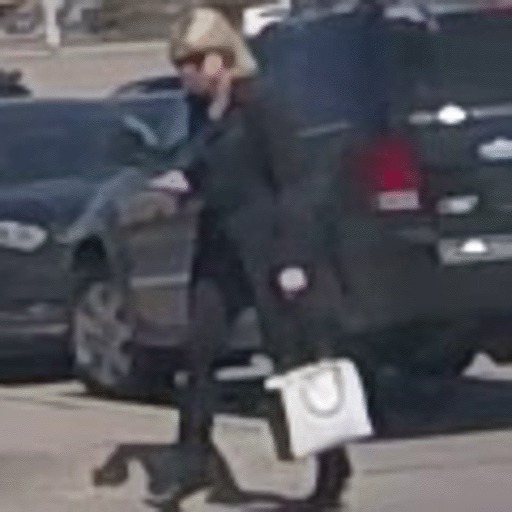}

}
\par\end{centering}
\noindent \begin{centering}
\subfloat[\label{fig:gesture}]{\includegraphics[scale=0.25]{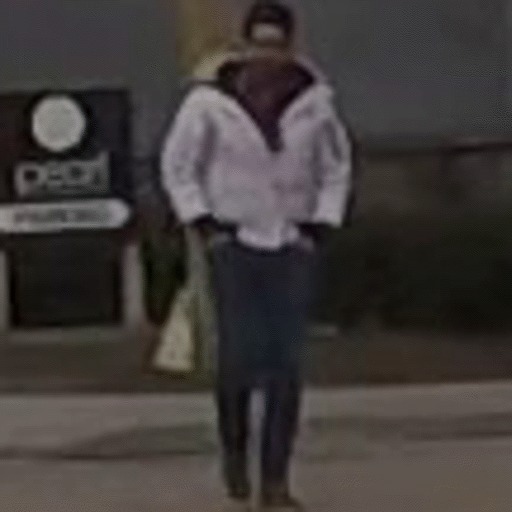}\includegraphics[scale=0.25]{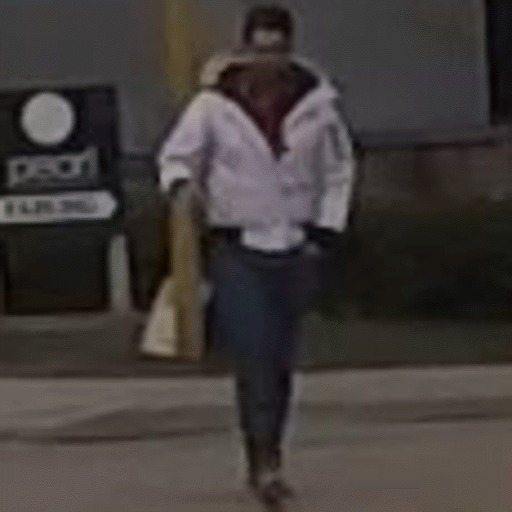}\includegraphics[scale=0.25]{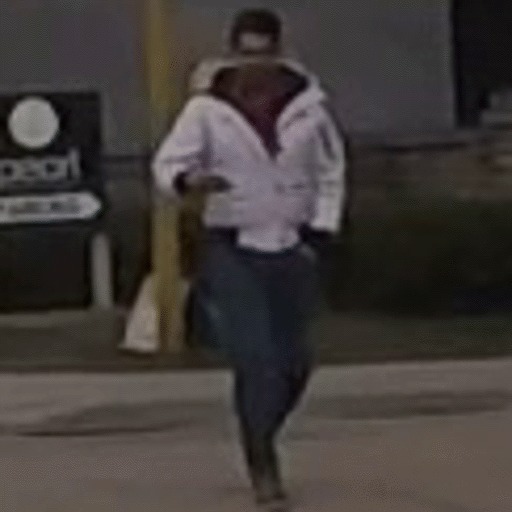}\includegraphics[scale=0.25]{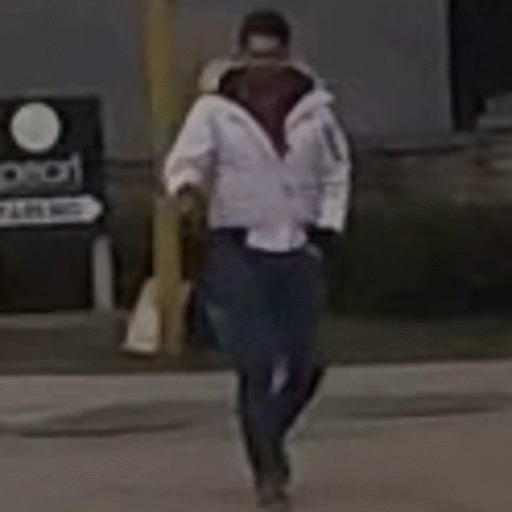}

}
\par\end{centering}
\noindent \begin{centering}
\subfloat[\label{fig:jaywalking}]{\includegraphics[scale=0.25]{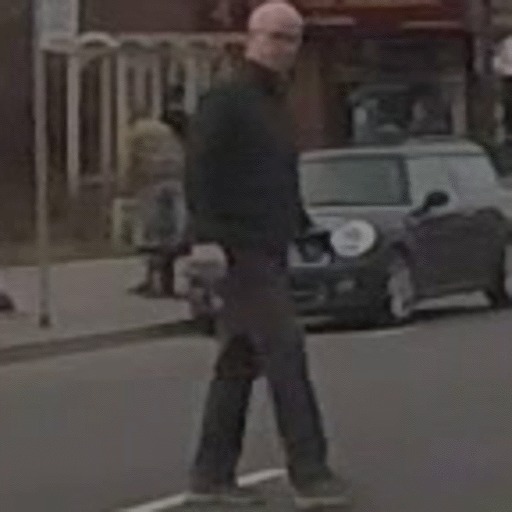}\includegraphics[scale=0.25]{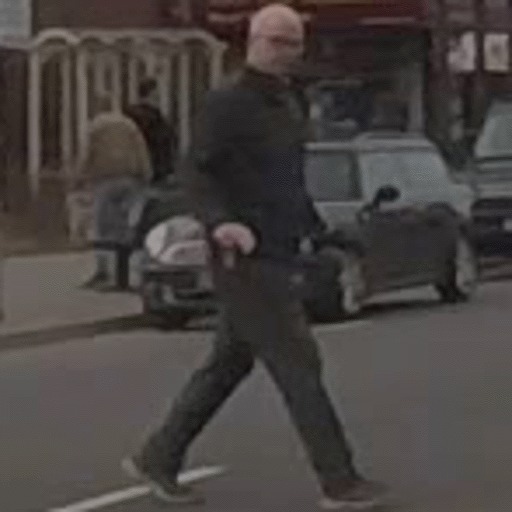}\includegraphics[scale=0.25]{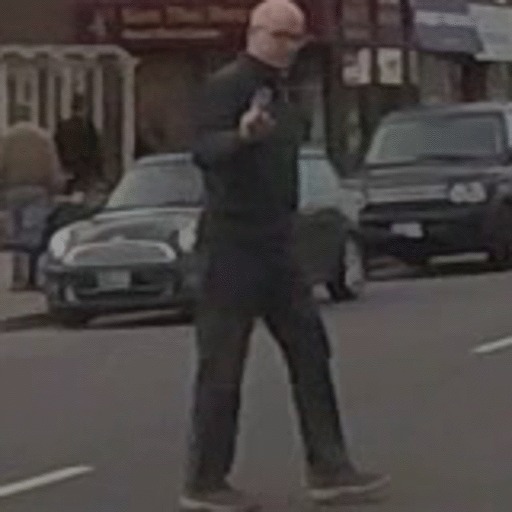}\includegraphics[scale=0.25]{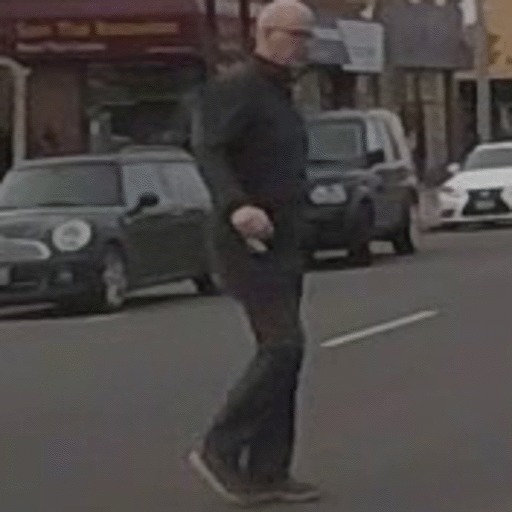}

}
\par\end{centering}
\noindent \begin{centering}
\subfloat[\label{fig:women_standing}]{\includegraphics[scale=0.25]{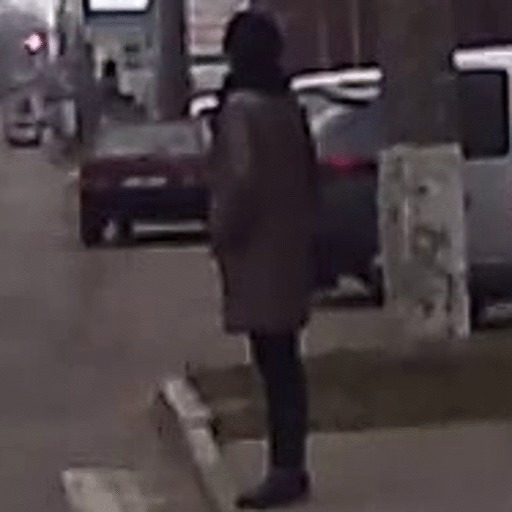}\includegraphics[scale=0.25]{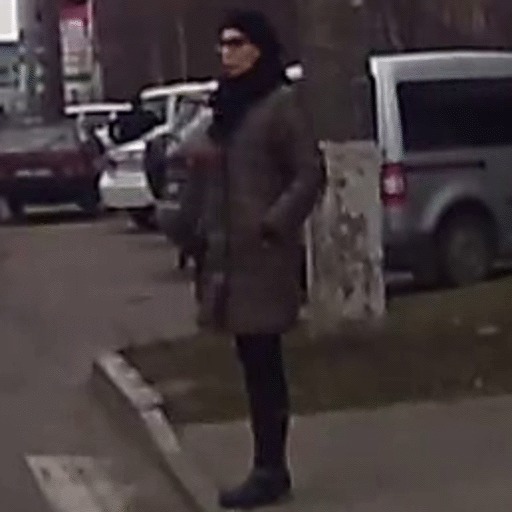}\includegraphics[scale=0.25]{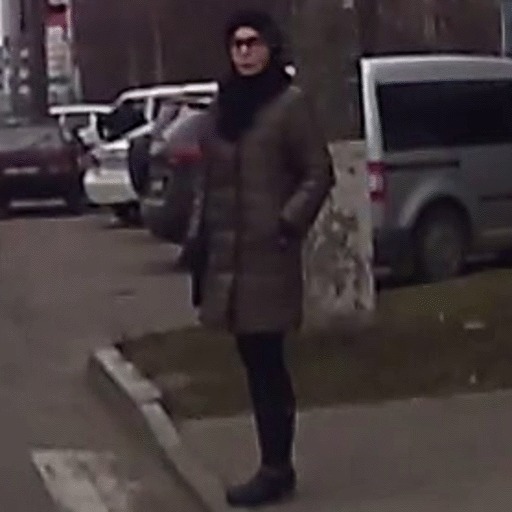}\includegraphics[scale=0.25]{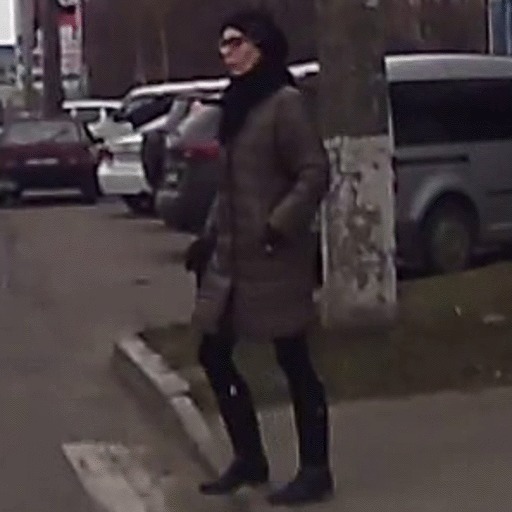}

}
\par\end{centering}
\caption{\label{fig:joint_attn}Examples of joint attention}
\end{figure*}

Surprisingly, despite increasing interest for joint attention in the
fields of robotics and human-robot interaction, it has not been explicitly
mentioned in the context of autonomous driving. For example, communication
between the driver and pedestrian is an instance of joint attention.
Consider the following scenario: a pedestrian crossing the street
(shown in \figref{woman_crossing}). Initially she is looking at her
cell phone, but as she approaches the curb, she looks up and slows
down because the vehicle is still moving. When the car slows down,
she speeds up and crosses the street. In this scenario all elements
of joint attention are apparent. Looking at the car and walking slower
is an observable attention behavior. The driver slowing down the car
indicates that he noticed the pedestrian and is yielding. His intention
is clearly interpreted as such, as the pedestrian speeds up and continues
to cross. A similar scene is shown in \figref{women_standing}. Here
the pedestrian is standing at the crossing and looking both ways to
find a gap in traffic. Again, once she notices that the driver is
slowing down, she begins to cross.

While these are fairly typical behaviors for marked crossings, there
are many more possible scenarios of communication between the traffic
participants. Humans recognize a myriad of ``social cues'' in everyday
traffic situations. Apart from establishing eye contact or waving
hands, people may be making assumptions about the way a driver would
behave based on visual characteristics such as the car\textquoteright s
make and model \cite{Gomes2014}. Understanding these social cues is
not always straightforward. Aside from visual processing challenges
such as variation in lighting conditions, weather or scene clutter,
there is also a need to understand the context in which the social
cue is observed. For instance, if the autonomous car sees someone
waving his hand, it needs to know whether it is a policeman directing
traffic, a pedestrian attempting to cross the street or someone hailing
a taxi. Consider \figref{gesture}, where a man is crossing the street
and makes a slight gesture as a signal of yielding to the driver,
or \figref{jaywalking}, where a man is jaywalking and acknowledges
the driver with a hand gesture. Responding to each of these scenarios
from the driver's perspective can be quite different and would require
high-level reasoning and deep scene analysis.

Today, automotive industry giants such as BMW, Tesla, Ford and Volkswagen,
who are actively working on autonomous driving systems, rely on visual
analysis technologies developed by Mobileye\footnote{\url{http://www.mobileye.com/}}
to handle obstacle avoidance, pedestrian detection or traffic scene
understanding. Mobileye's approach to solving visual tasks is to use
deep learning techniques which require a large amount of data collected
from hundreds of hours of driving. This system has been successfully
tested and is currently being used in semi-autonomous vehicles. However,
the question remains open whether deep learning suffices for achieving
full autonomy in which tasks are not limited to detection of pedestrians,
cars or obstacles (which are not still fully reliable \cite{Kelion2016,Thielman2016}),
but also involve merging with ongoing traffic, dealing with unexpected
behaviors such as jaywalking, responding to emergency vehicles, and
yielding to other vehicles or pedestrians at intersections.

To answer this question we need to consider the following characteristics
of deep learning algorithms. First, even though deep learning algorithms
perform very well in tasks such as object recognition, they lack the
ability to establish causal relationships between what is observed
and the context in which it has occurred \cite{Vincent2016, Knight2015a}.
This problem also has been empirically demonstrated by training neural
networks over various types of data \cite{Vincent2016}.

The second limitation of deep learning is the lack of robustness to
changes in visual input \cite{Goertzel2015}. This problem can occur
when a deep neural network misclassifies an object due to minor changes
(at a pixel level) to an image \cite{Szegedy2014} or even recognizes
an object from a randomly generated image \cite{Nguyen2015}.

\section{Existing datasets}

The autonomous driving datasets currently available to the public
are primarily intended for applications such as 3D mapping, navigation,
and car and pedestrian detection. Out of these datasets only a limited
number contain data that can be used for behavioral studies. Below
some of these datasets are listed.
\begin{itemize}
\item KITTI \cite{KITTI2016b}: This is perhaps one of the most known publicly
available datasets for autonomous driving. It contains data collected
from various locations such as residential areas and city streets,
highways and gated environments. The main application is for 3D reconstruction,
3D detection, tracking and visual odometry. Some of the videos in
KITTI show pedestrians, other vehicles and cyclists movements alongside
the car. The data has no annotation of their behaviors.

\item Caltech pedestrian detection benchmark \cite{Dollar2012}: This is
a very large dataset of pedestrians consisting of approximately 10 hours of driving in regular traffic in urban environments. The annotations include temporal correspondence between bounding boxes around pedestrians and detailed occlusion lables.

\item Berkeley pedestrian dataset \cite{Fragkiadaki2012}: This dataset consists
of a large number of videos of pedestrians collected from a stationary
car at street intersections. Bounding boxes around pedestrians are
provided for pedestrian detection and tracking.
\item Semantic Structure From Motion (SSFM) \cite{Bao2012}: As the name
implies, this dataset is collected for scene understanding. The annotation
is limited to bounding boxes around the objects of interest and name
tags for the purpose of detection. This dataset includes a number
of street view videos of cars and pedestrians walking.
\item The German Traffic Sign Detection Benchmark \cite{Sign2015a}: This
dataset consists of 900 high-resolution images of roads and streets
some of which show pedestrians crossing and cars. The ground truth
for the dataset only specifies the positions of traffic signs in the
images.
\item The .enpeda.. (environment perception and driver assistance) Image
Sequence Analysis Test Site (EISATS) \cite{Klette2014}: EISATS contains
short synthetic and real videos of cars driving on roads and streets.
The sole purpose of this dataset is comparative performance evaluation
of stereo vision and motion analysis. The available annotation is
limited to the camera's intrinsic parameters.
\item Daimler Pedestrian Benchmark Datasets \cite{Gavril2015}: These are
particularly useful datasets for various scenarios of pedestrian detection
such as segmentation, classification, and path prediction. The sensors
of choice are monocular and binocular cameras and the datasets contain
both color and grayscale images. The ground truth data is limited
to the detection applications and does not include any behavioral
analysis.
\item UvA Person Tracking from Overlapping Cameras Datasets  \cite{Gavril2015a}:
These datasets mainly are concerned with the tasks of tracking, and
pose and trajectory estimation using multiple cameras. The ground
truth is also limited to only facilitate tracking applications.
\end{itemize}
In recent years traffic behavior of drivers and pedestrians has became
a widely studied topic for collision prevention and traffic safety.
Several large-scale naturalistic driving studies have been conducted
in the USA \cite{100car,110Car,SHRP2}, which accumulated over 4 petabytes
of data (video, audio, instrumental, traffic, weather, etc) from hundreds
of volunteer drivers in multiple locations. However, only some depersonalized
general statistics are available to the general public \cite{VTTI},
while only qualified researchers have access to the raw video and
sensor data.

\section{The JAAD Dataset}

The JAAD dataset was created to facilitate studying the behavior of
traffic participants. The data consists of 346 high-resolution video
clips (5-15s) with annotations showing various situations typical
for urban driving. These clips were extracted from approx. 240 hours
of driving videos collected in several locations. Two vehicles equipped
with wide-angle video cameras were used for data collection (\tabref{dataset}).
Cameras were mounted inside the cars in the center of the windshield
below the rear view mirror.
\begin{figure*}
\noindent \begin{centering}
\includegraphics[scale=0.7]{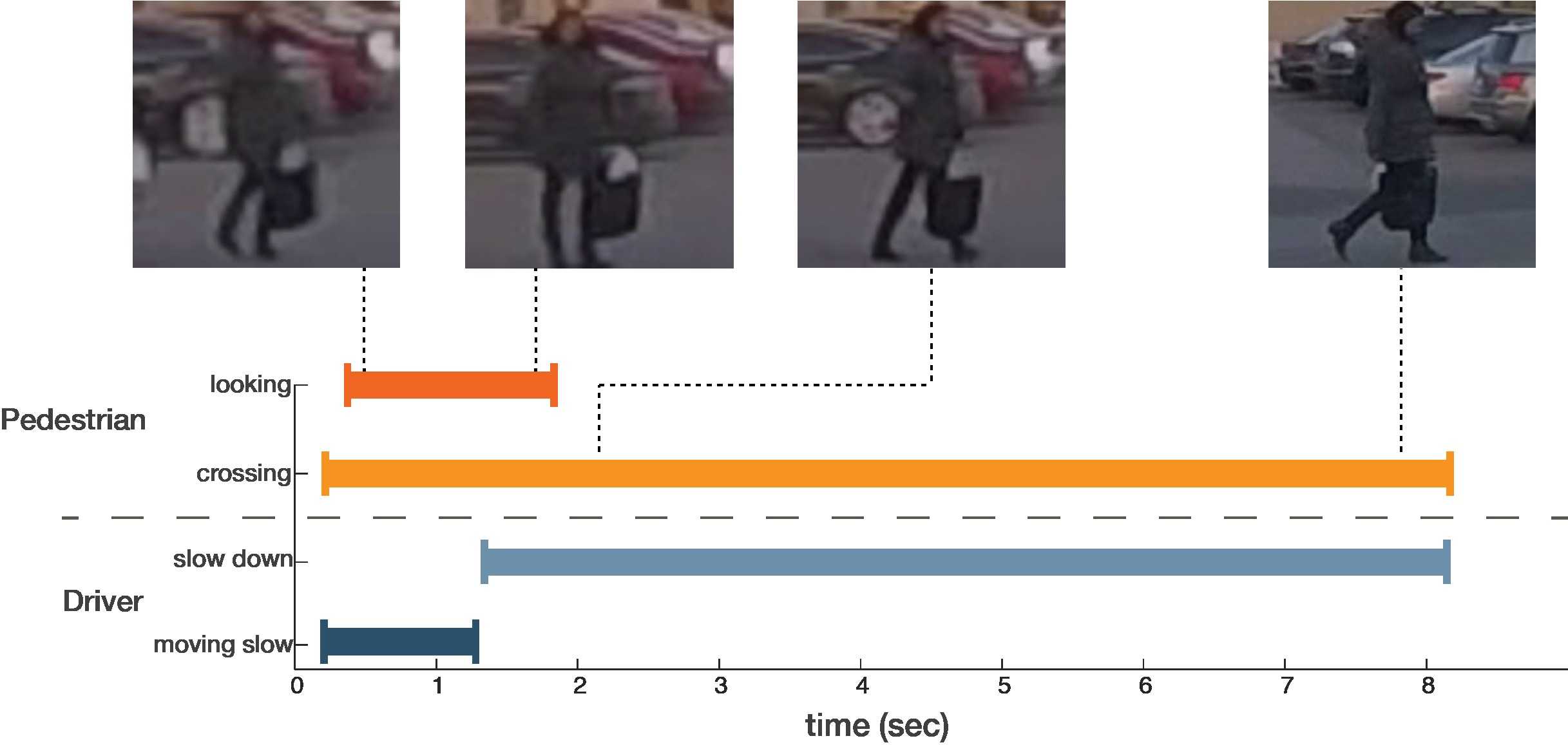}
\par\end{centering}
\caption{\label{fig:timeline_of_crossing} A timeline of events recovered from the behavioral data. Here a single pedestrian is crossing the parking
lot. Initially the driver is moving slow and, as he notices the pedestrian
ahead, slows down to let her pass. At the same time the pedestrian
crosses without looking first, then turns to check if the road is
safe, and as she sees the driver yielding, continues to cross. The
difference in resolution between the images is due to the changes
in distance to the pedestrian as the car moves forward.}
\end{figure*}

\begin{figure*}
\noindent \begin{centering}
\subfloat[]{\includegraphics[scale=0.25]{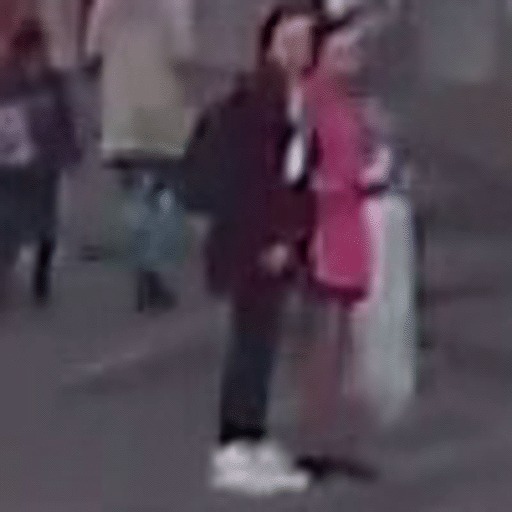}\includegraphics[scale=0.25]{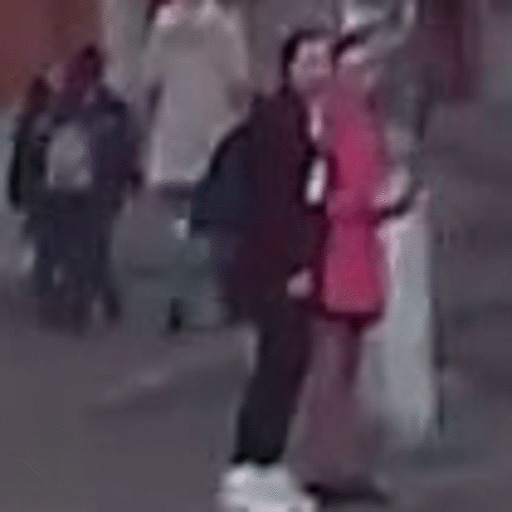}\includegraphics[scale=0.25]{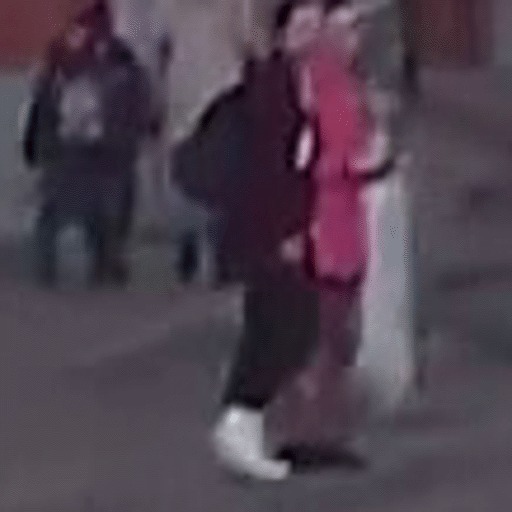}\includegraphics[scale=0.25]{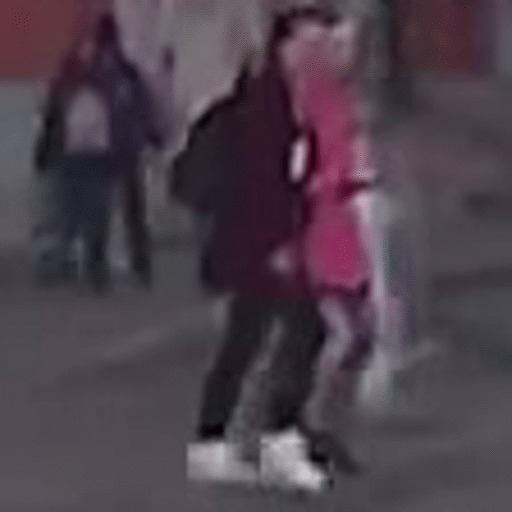}

}
\par\end{centering}
\noindent \begin{centering}
\subfloat[]{\includegraphics[scale=0.25]{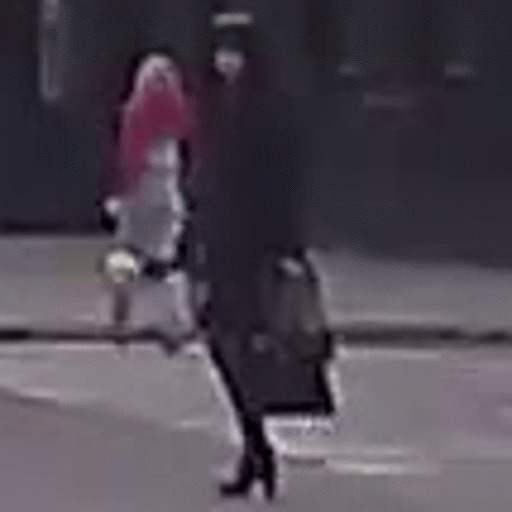}\includegraphics[scale=0.25]{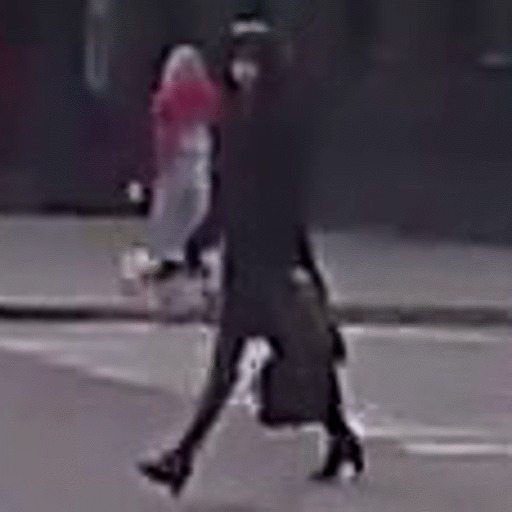}\includegraphics[scale=0.25]{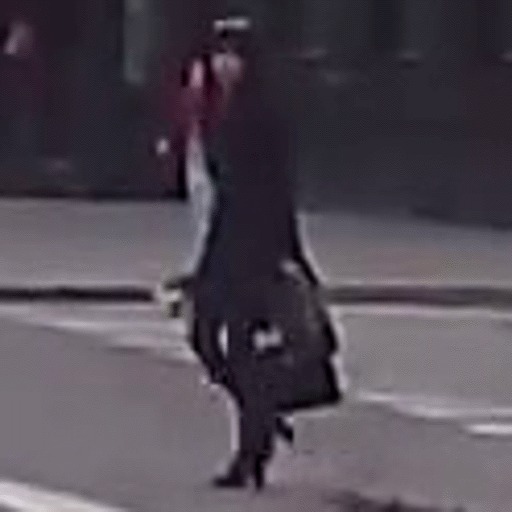}\includegraphics[scale=0.25]{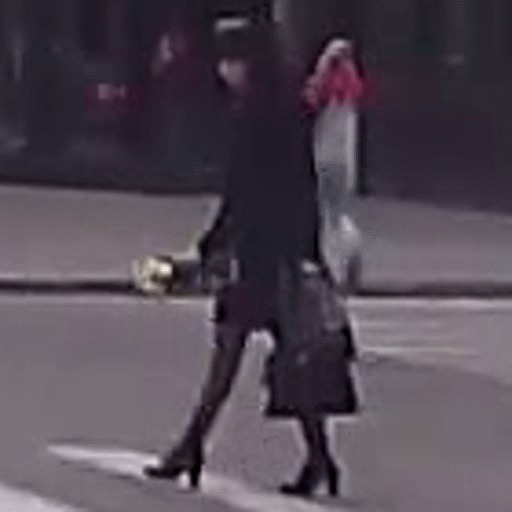}

}
\par\end{centering}
\noindent \begin{centering}
\subfloat[\label{fig:look-back}]{\includegraphics[scale=0.25]{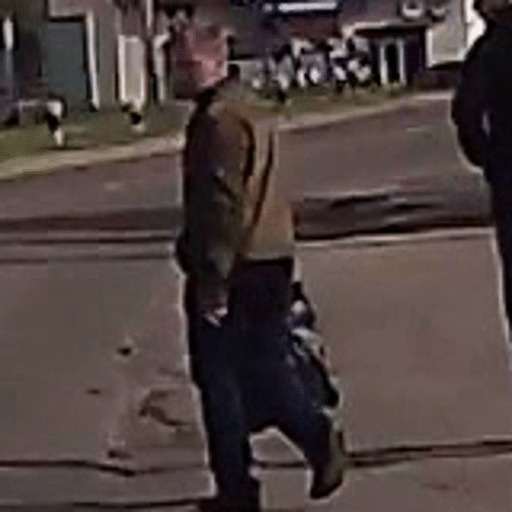}\includegraphics[scale=0.25]{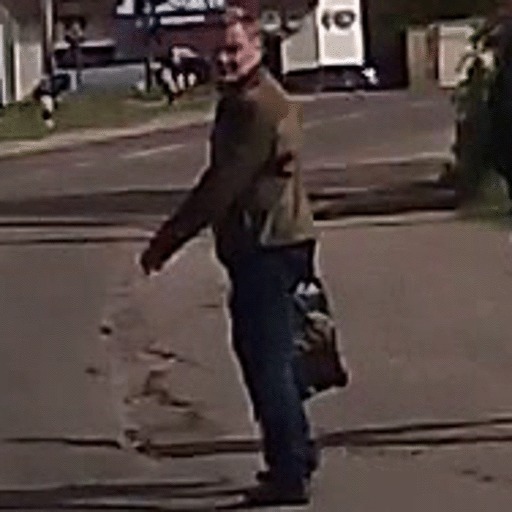}\includegraphics[scale=0.25]{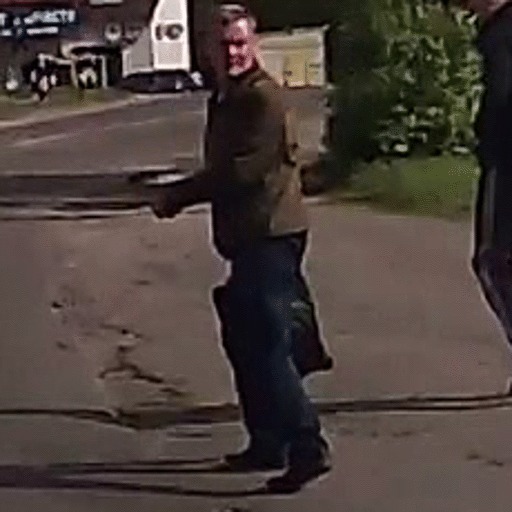}\includegraphics[scale=0.25]{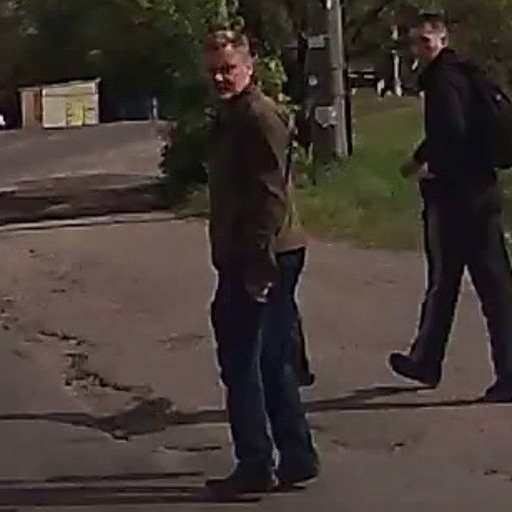}

}
\par\end{centering}
\noindent \begin{centering}
\subfloat[\label{fig:peeking}]{\includegraphics[scale=0.25]{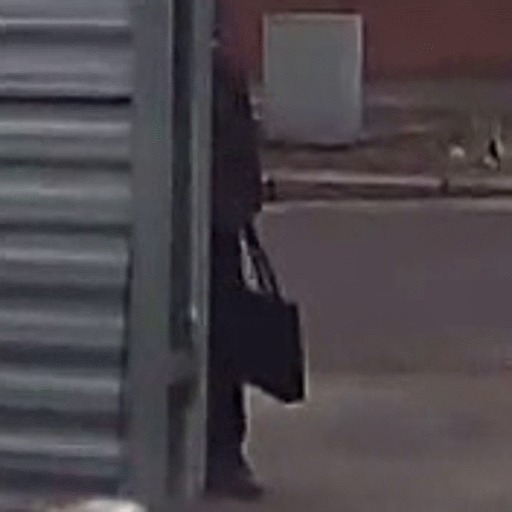}\includegraphics[scale=0.25]{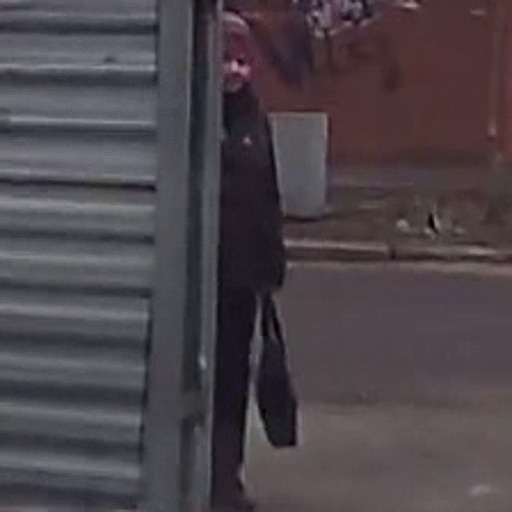}\includegraphics[scale=0.25]{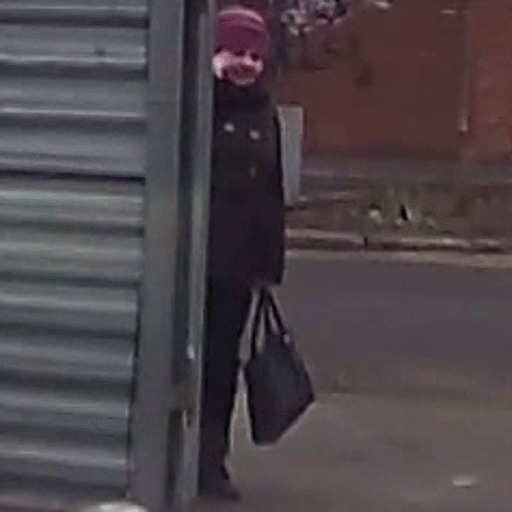}\includegraphics[scale=0.25]{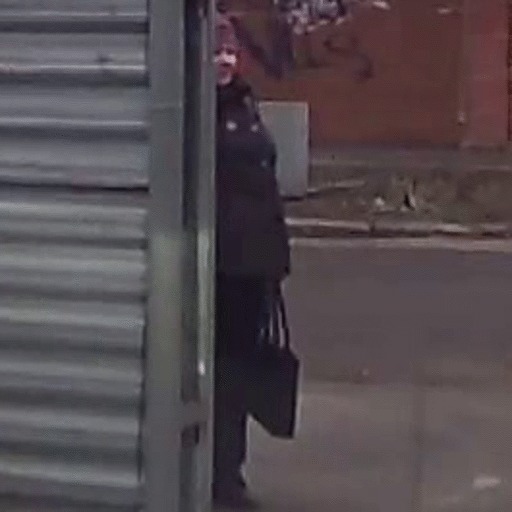}

}
\par\end{centering}
\caption{\label{fig:joint_attn-1}More examples of joint attention}
\end{figure*}

The video clips represent a wide variety of scenarios involving pedestrians
and other drivers. Most of the data is collected in urban areas (downtown
and suburban), only a few clips are filmed in rural locations. Many
of the situations resemble the ones we have described earlier, where
pedestrians wait at the designated crossings. In other samples pedestrians
may be walking along the road and look back to see if there is a gap
in traffic (\figref{look-back}), peek from behind the obstacle to
see if it is safe to cross \figref{peeking}, waiting to cross on a
divider between the lanes, carrying heavy objects or walking with
children or pets. Our dataset captures pedestrians of various ages
walking alone and in groups, which may be a factor affecting their
behavior. For example, elderly people and
\begin{table}[h]
\noindent \begin{centering}
\resizebox{\columnwidth}{!}{
\begin{tabular*}{1\columnwidth}{@{\extracolsep{\fill}}c|c|c|>{\centering}p{0.25\columnwidth}}
{\footnotesize{}\# of clips} & {\footnotesize{}Location} & {\footnotesize{}Resolution} & {\footnotesize{}Camera model}\tabularnewline
\hline 
\hline 
{\footnotesize{}55} & {\footnotesize{}North York, ON, Canada} & {\footnotesize{}$1920\times1080$} & {\footnotesize{}GoPro HERO+}\tabularnewline
\hline 
{\footnotesize{}276} & {\footnotesize{}Kremenchuk, Ukraine} & {\footnotesize{}$1280\times720$} & {\scriptsize{}Highscreen Black Box Connect}\tabularnewline
\hline 
{\footnotesize{}6} & {\footnotesize{}Hamburg, Germany} & {\footnotesize{}$1280\times720$} & {\scriptsize{}Highscreen Black Box Connect}\tabularnewline
\hline 
{\footnotesize{}5} & {\footnotesize{}New York, USA} & {\footnotesize{}$1920\times1080$} & {\footnotesize{}GoPro HERO+}\tabularnewline
\hline 
{\footnotesize{}4} & {\footnotesize{}Lviv, Ukraine} & {\footnotesize{}$1920\times1080$} & {\footnotesize{}Garmin GDR-35}\tabularnewline
\end{tabular*}
}
\par\end{centering}
\caption{\label{tab:dataset}Locations and equipment used to capture videos
in the JAAD dataset}
\end{table}
parents with children may walk slower and be more cautious. The dataset
contains fewer clips of interactions with other drivers, most of them
occur in uncontrolled intersections, in parking lots or when another
driver is moving across several lanes to make a turn.

\begin{figure*}
\noindent \begin{centering}
\subfloat[Sunset]{\includegraphics[scale=0.13]{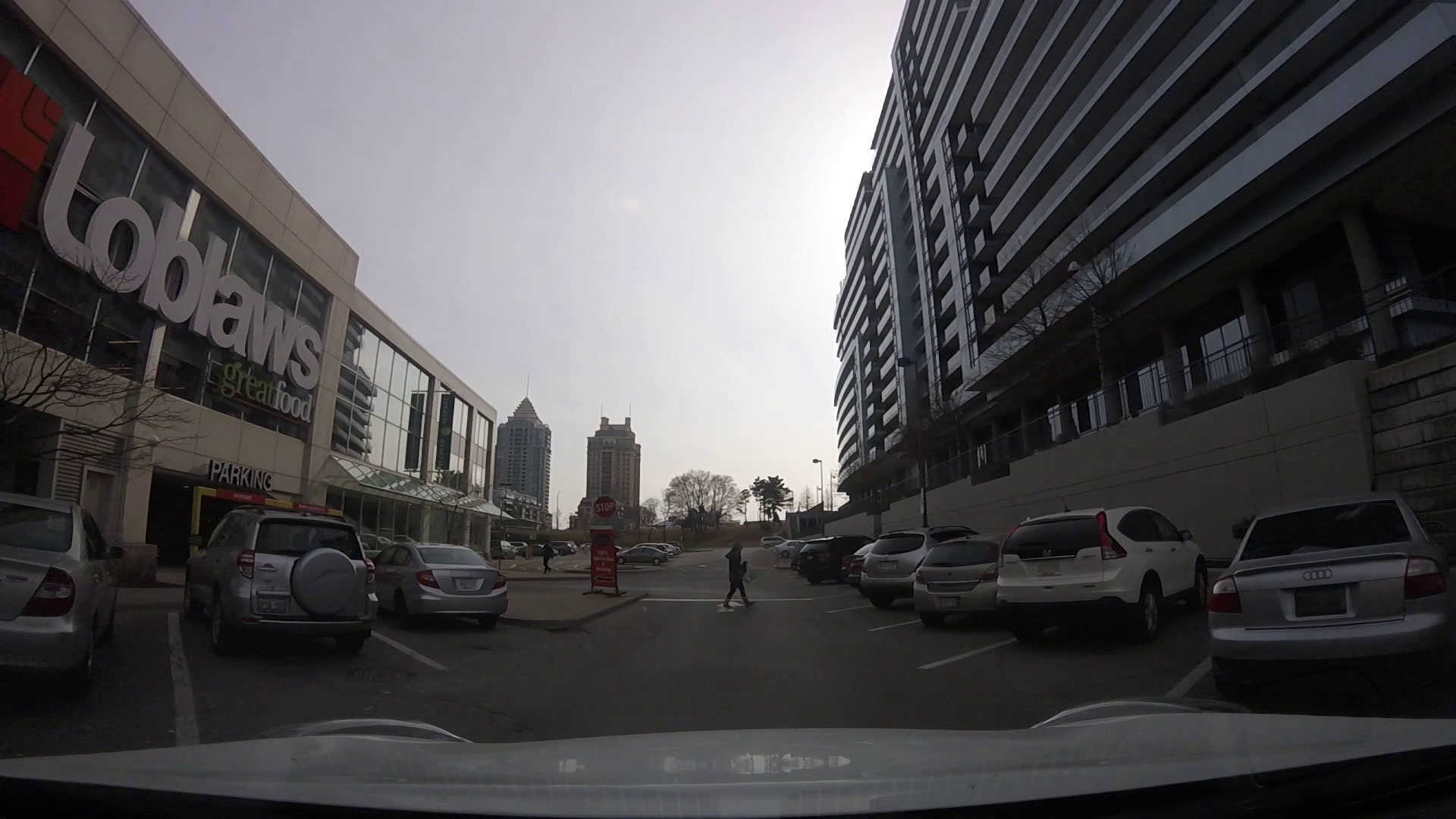}

}\subfloat[\label{fig:Sunrise}Sunrise]{\includegraphics[scale=0.13]{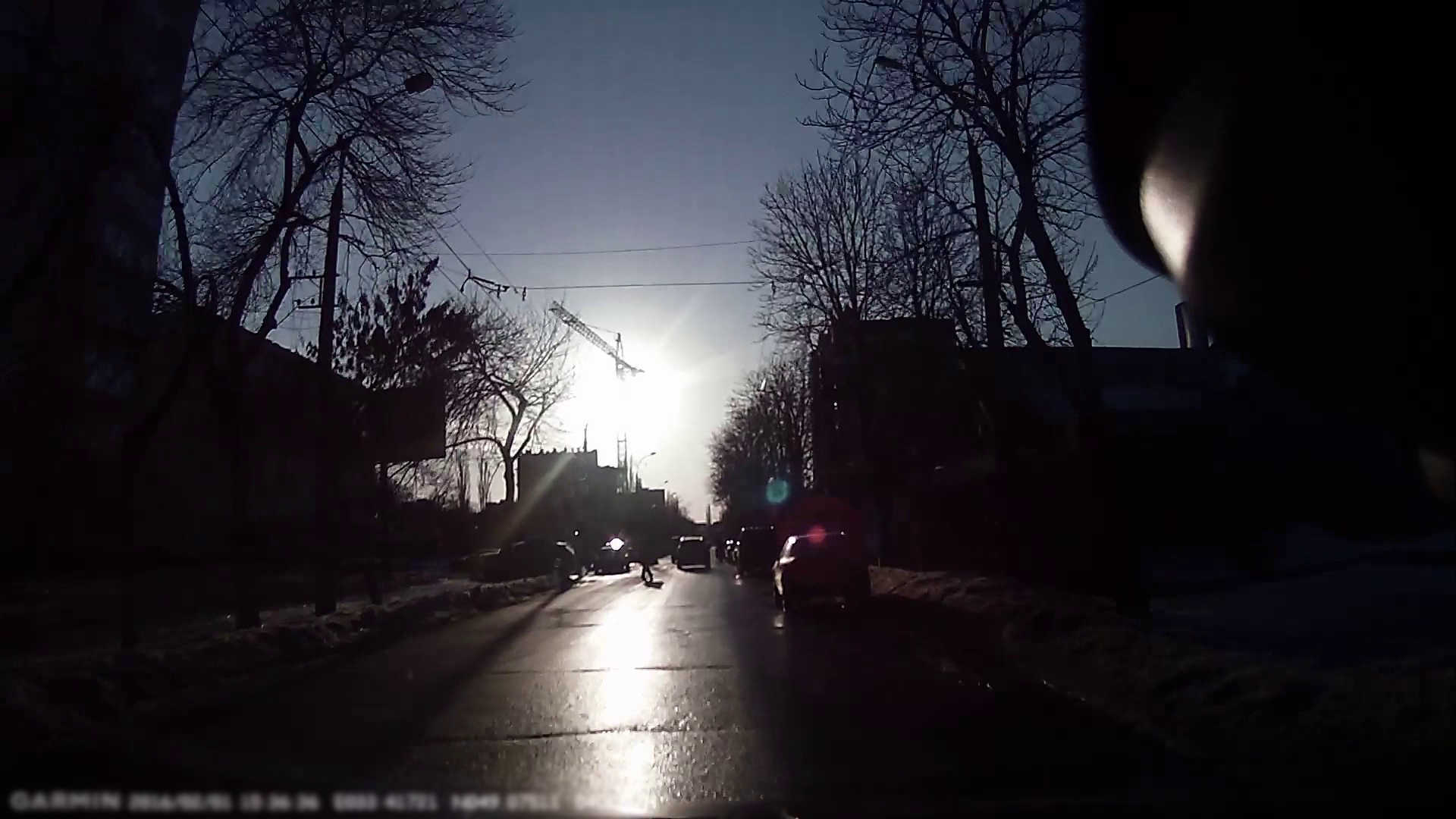}

}
\par\end{centering}
\noindent \begin{centering}
\subfloat[After a heavy snowfall]{\includegraphics[scale=0.13]{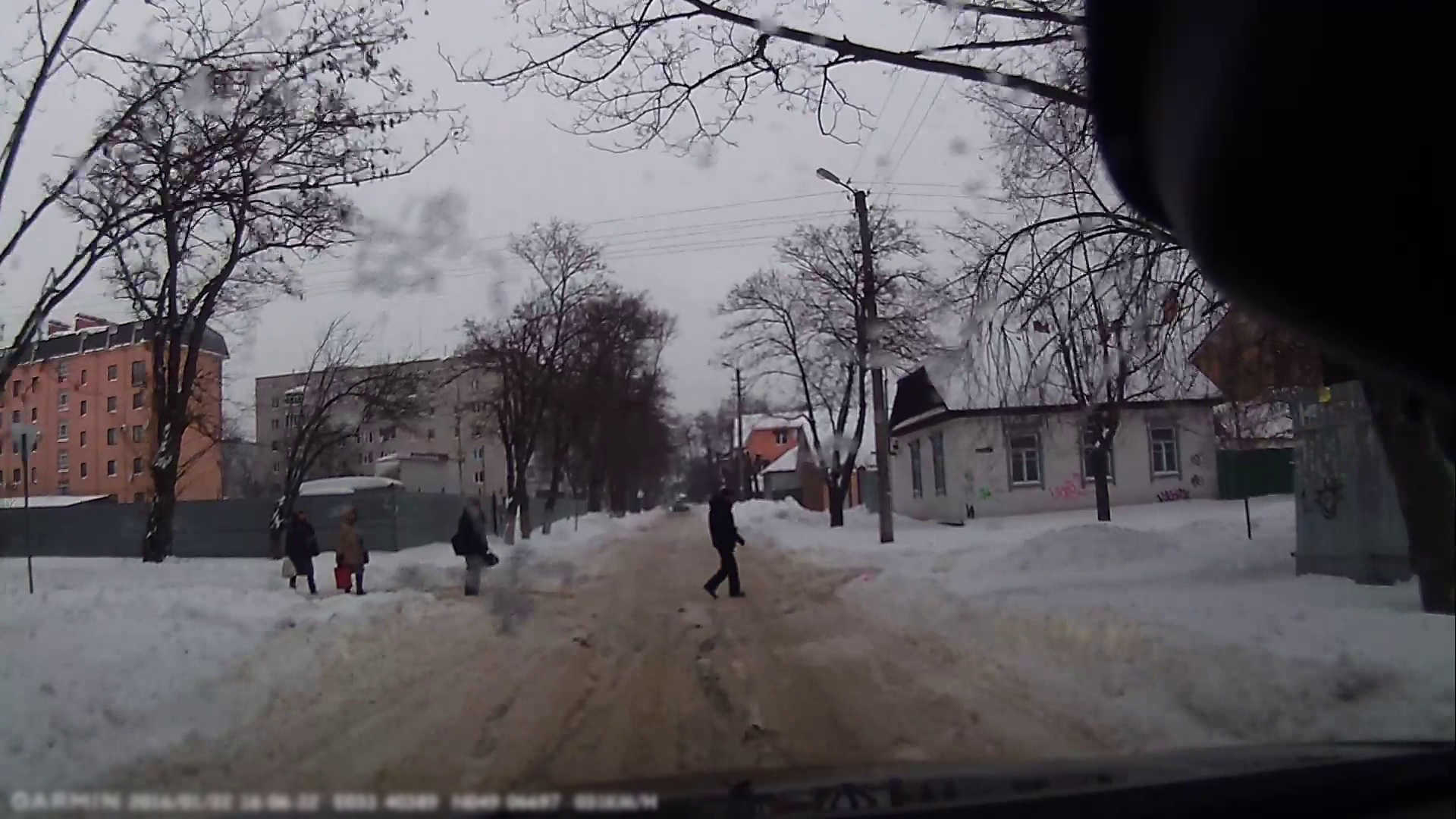}

}\subfloat[During a heavy rain]{\includegraphics[scale=0.13]{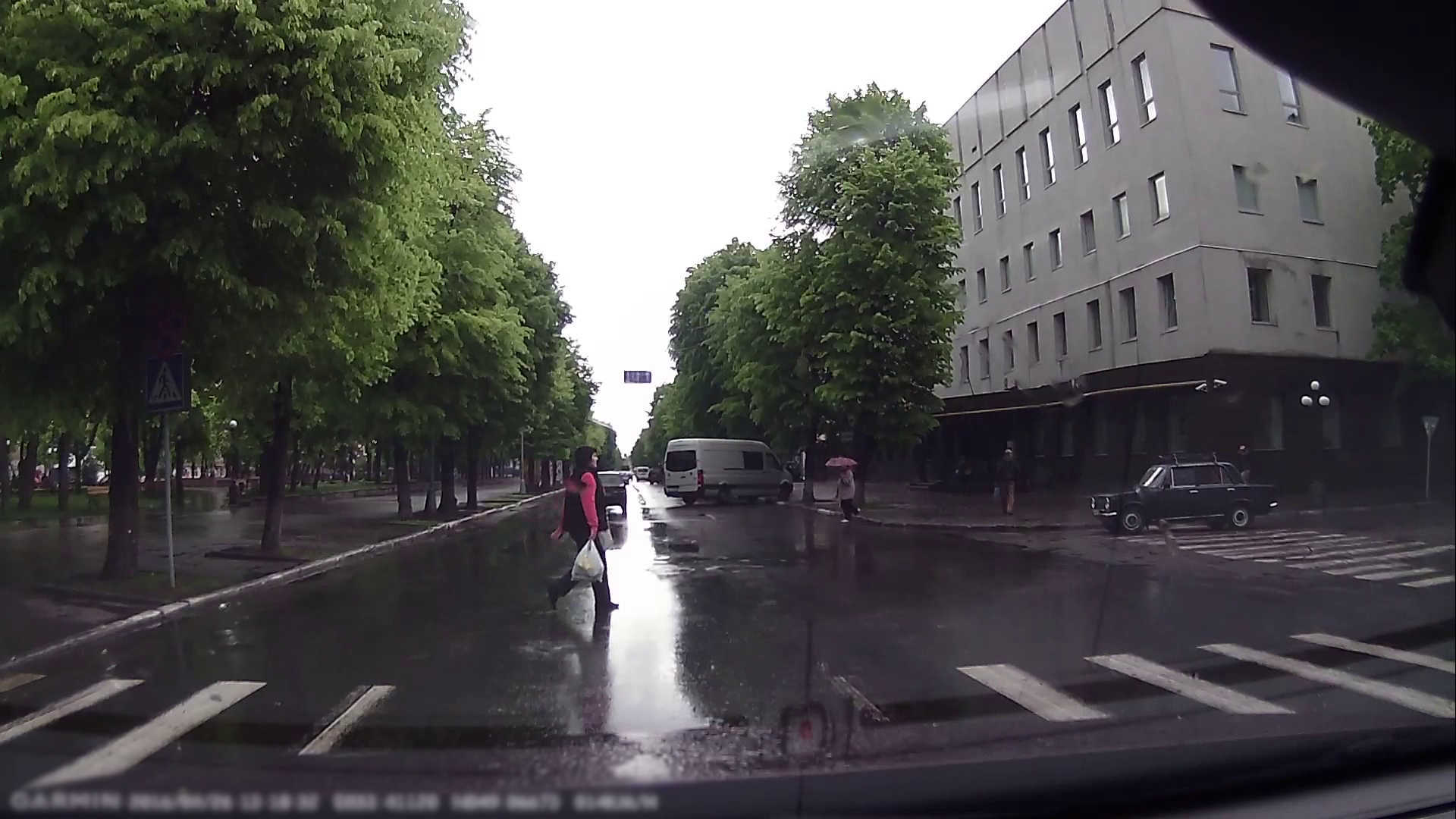}

}
\par\end{centering}
\noindent \begin{centering}
\subfloat[Multiple pedestrians crossing]{\includegraphics[scale=0.13]{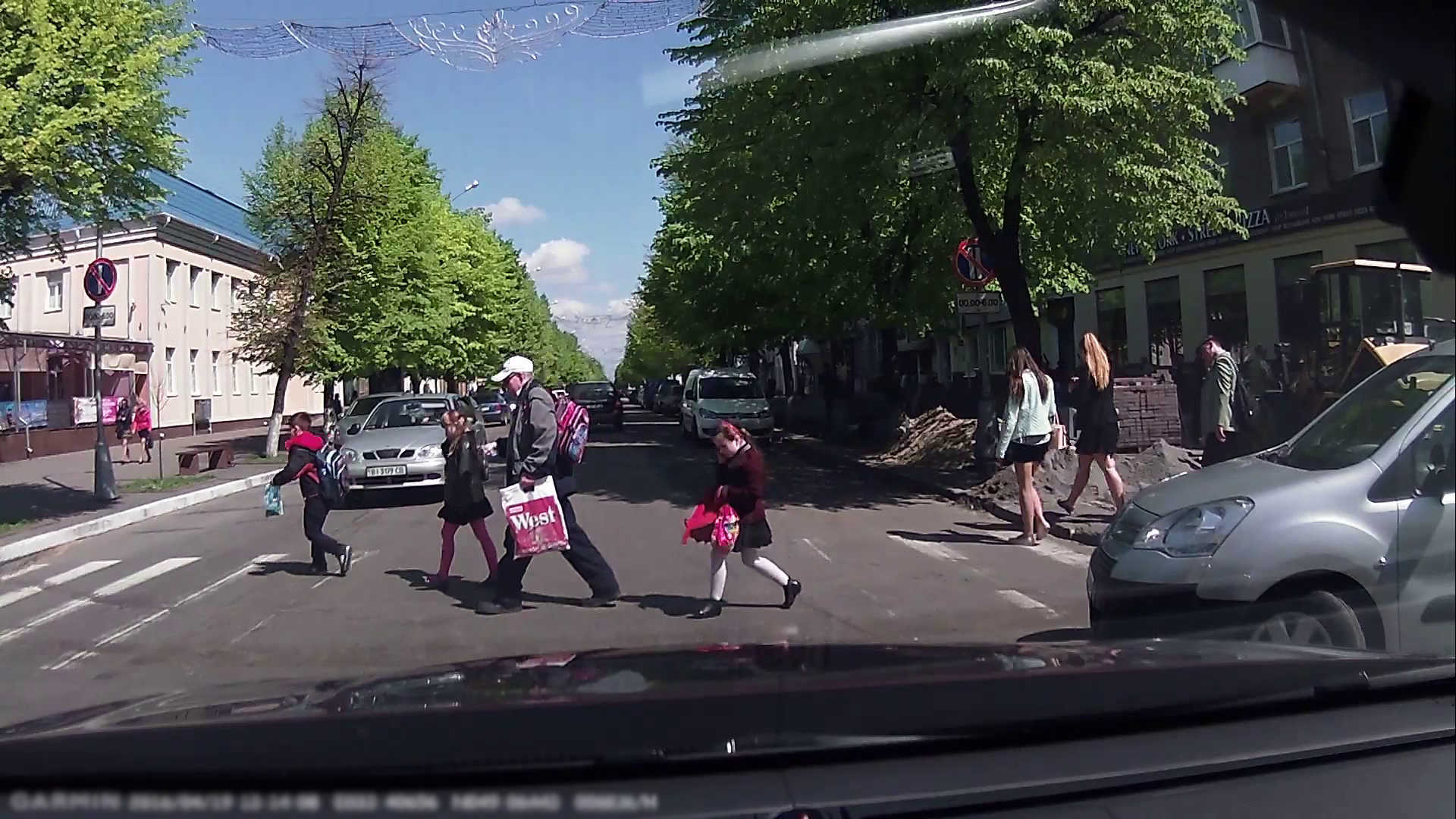}

}\subfloat[At the parking lot]{\includegraphics[scale=0.13]{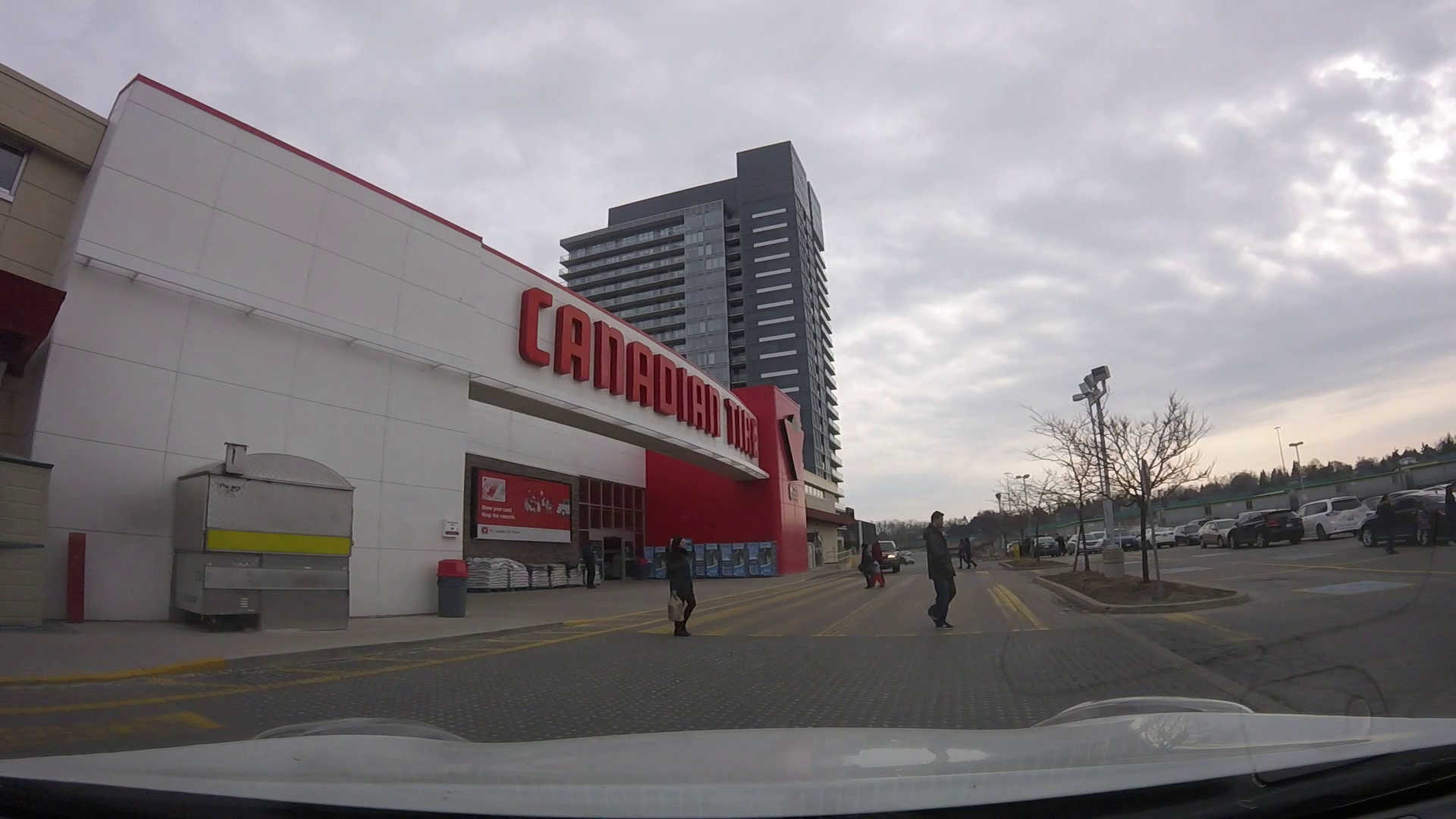}

}
\par\end{centering}
\caption{\label{fig:Sample-frames}Sample frames from the dataset showing different
weather conditions and locations.}

\end{figure*}

Most of the videos in the dataset were recorded during the daytime
and only a few clips were filmed at night, sunset and sunrise. The
last two conditions are particularly challenging, as the sun is glaring
directly into the camera (\figref{Sunrise}). We also tried to capture
a variety of weather conditions (\figref{Sample-frames}), as yet
another factor affecting the behavior of traffic participants. For
example, during the heavy snow or rain people wearing hooded jackets
or carrying umbrellas may have limited visibility of the road. Since
their faces are obstructed it is also harder to tell if they are paying
attention to the traffic from the driver's perspective.

We attempted to capture all of these conditions for further analysis
by providing two kinds of annotations for the data: bounding boxes
and textual annotations. Bounding boxes are provided only for cars
and pedestrians that interact with or require attention of the driver
(e.g. another car yielding to the driver, pedestrian waiting to cross
the street, etc.). Bounding boxes for each video are written into
an xml file with frame number, coordinates, width, height and occlusion
flag.
\begin{table*}
\begin{centering}
\begin{tabular}{c|c}
Categorical variable & Values\tabularnewline
\hline 
\hline 
time\_of\_day & day/night\tabularnewline
\hline 
weather & clear/snow/rain/cloudy\tabularnewline
\hline 
location & street/indoor/parking\_lot\tabularnewline
\hline 
designated\_crossing & yes/no\tabularnewline
\hline 
age\_gender & Child/Young/Adult/Senior Male/Female\tabularnewline
\end{tabular}\quad{}\quad{}\quad{}%
\begin{tabular}{c|c}
Behavior event & Type\tabularnewline
\hline 
\hline 
Crossing & state\tabularnewline
\hline 
Stopped & state\tabularnewline
\hline 
Moving fast & state\tabularnewline
\hline 
Moving slow & state\tabularnewline
\hline 
Speed up & state\tabularnewline
\hline 
Slow down & state\tabularnewline
\hline 
Clear path & state\tabularnewline
\hline 
Looking & state\tabularnewline
\hline 
Look & point\tabularnewline
\hline 
Signal & point\tabularnewline
\hline 
Handwave & point\tabularnewline
\end{tabular}
\par\end{centering}
\caption{\label{tab:Behavioral-data}Variables associated with each video and types of events represented in the dataset. There are two types of
behavior events: state and point. State event may have an arbitrary
duration, while point events last a short fixed amount of time (0.1
sec) and signify a quick glance or gestures made by pedestrians.}
\end{table*}
\begin{figure*}
\textsf{\scriptsize{}\qquad{}\qquad{}\qquad{}Observation id	GOPR0103\_528\_542}{\scriptsize \par}

\textsf{\scriptsize{}\qquad{}\qquad{}\qquad{}Media file(s)}{\scriptsize \par}

\textsf{\scriptsize{}\qquad{}\qquad{}\qquad{}Player \#1	GOPR0103\_528\_542.MP4}{\scriptsize \par}

\textsf{\scriptsize{}\qquad{}\qquad{}\qquad{}Observation date	2016-07-15
15:15:38}{\scriptsize \par}

\textsf{\scriptsize{}\qquad{}\qquad{}\qquad{}Description}{\scriptsize \par}

\textsf{\scriptsize{}\qquad{}\qquad{}\qquad{}Time offset (s)	0.000}{\scriptsize \par}

\begin{raggedright}
\textsf{\scriptsize{}\qquad{}\qquad{}\qquad{}Independent variables}
\par\end{raggedright}{\scriptsize \par}
\begin{raggedright}
\textsf{\scriptsize{}\qquad{}\qquad{}\qquad{}}%
\begin{tabular}{lc}
\textsf{\scriptsize{}variable} & \textsf{\scriptsize{}value}\tabularnewline
\hline 
\textsf{\scriptsize{}weather} & \textsf{\scriptsize{}rain}\tabularnewline
\textsf{\scriptsize{}age\_gender} & \textsf{\scriptsize{}AF}\tabularnewline
\textsf{\scriptsize{}designated} & \textsf{\scriptsize{}no}\tabularnewline
\textsf{\scriptsize{}location} & \textsf{\scriptsize{}plaza}\tabularnewline
\textsf{\scriptsize{}time\_of\_day} & \textsf{\scriptsize{}daytime}\tabularnewline
\end{tabular}
\par\end{raggedright}{\scriptsize \par}
\noindent \begin{raggedright}
\textsf{\scriptsize{}\qquad{}\qquad{}}%
\begin{tabular}{lccccccc}
\textsf{\scriptsize{}Time} & \textsf{\scriptsize{}Media file path} & \textsf{\scriptsize{}Media total length} & \textsf{\scriptsize{}FPS} & \textsf{\scriptsize{}Subject} & \textsf{\scriptsize{}Behavior} & \textsf{\scriptsize{}Comment} & \textsf{\scriptsize{}Status}\tabularnewline
\hline 
\textsf{\scriptsize{}0.19} & \textsf{\scriptsize{}GOPR0088\_335\_344.MP4} & \textsf{\scriptsize{}9.01} & \textsf{\scriptsize{}29.97} & \textsf{\scriptsize{}Driver} & \textsf{\scriptsize{}moving slow} &  & \textsf{\scriptsize{}START}\tabularnewline
\textsf{\scriptsize{}0.208} & \textsf{\scriptsize{}GOPR0088\_335\_344.MP4} & \textsf{\scriptsize{}9.01} & \textsf{\scriptsize{}29.97} & \textsf{\scriptsize{}pedestrian} & \textsf{\scriptsize{}crossing} &  & \textsf{\scriptsize{}START}\tabularnewline
\textsf{\scriptsize{}0.308} & \textsf{\scriptsize{}GOPR0088\_335\_344.MP4} & \textsf{\scriptsize{}9.01} & \textsf{\scriptsize{}29.97} & \textsf{\scriptsize{}pedestrian} & \textsf{\scriptsize{}looking} &  & \textsf{\scriptsize{}START}\tabularnewline
\textsf{\scriptsize{}1.301} & \textsf{\scriptsize{}GOPR0088\_335\_344.MP4} & \textsf{\scriptsize{}9.01} & \textsf{\scriptsize{}29.97} & \textsf{\scriptsize{}Driver} & \textsf{\scriptsize{}moving slow} &  & \textsf{\scriptsize{}STOP}\tabularnewline
\textsf{\scriptsize{}1.302} & \textsf{\scriptsize{}GOPR0088\_335\_344.MP4} & \textsf{\scriptsize{}9.01} & \textsf{\scriptsize{}29.97} & \textsf{\scriptsize{}Driver} & \textsf{\scriptsize{}slow down} &  & \textsf{\scriptsize{}START}\tabularnewline
\textsf{\scriptsize{}1.892} & \textsf{\scriptsize{}GOPR0088\_335\_344.MP4} & \textsf{\scriptsize{}9.01} & \textsf{\scriptsize{}29.97} & \textsf{\scriptsize{}pedestrian} & \textsf{\scriptsize{}looking} &  & \textsf{\scriptsize{}STOP}\tabularnewline
\textsf{\scriptsize{}8.351} & \textsf{\scriptsize{}GOPR0088\_335\_344.MP4} & \textsf{\scriptsize{}9.01} & \textsf{\scriptsize{}29.97} & \textsf{\scriptsize{}pedestrian} & \textsf{\scriptsize{}crossing} &  & \textsf{\scriptsize{}STOP}\tabularnewline
\textsf{\scriptsize{}8.99} & \textsf{\scriptsize{}GOPR0088\_335\_344.MP4} & \textsf{\scriptsize{}9.01} & \textsf{\scriptsize{}29.97} & \textsf{\scriptsize{}Driver} & \textsf{\scriptsize{}slow down} &  & \textsf{\scriptsize{}STOP}\tabularnewline
\end{tabular}
\par\end{raggedright}{\scriptsize \par}
\caption{\label{fig:Textual-annotation-example}Example of textual annotation for a video created using BORIS. The file contains the id and the name of the video file, a tab-separated list of independent variables
(weather, age and gender of pedestrians, whether the crossing is designated
or not, location and time of the day) and a tab-separated list of
events. Each event has an associated time stamp, subject, behavior
and status, which may be used to recover sequence of events for analysis.}
\end{figure*}

Textual annotations are created using BORIS\footnote{\url{http://www.boris.unito.it/}} \cite{friard2016boris}
- event logging software for video observations. It allows to assign
predefined behaviors to different subjects seen in the video, and
can also save some additional data, such as video file id, location
where the observation has been made, etc.

A list of all behaviors, independent variables and their values is
shown in \tabref{Behavioral-data}. We save the following data for
each video clip: weather, time of the day, age and gender of the pedestrians,
location and whether it is a designated crosswalk. Each pedestrian
is assigned a label (pedestrian1, pedestrian2, etc.). We also distinguish
between the driver inside the car and other drivers, which are labeled
as ``Driver'' and ``Driver\_other'' respectively. This is necessary
for the situations where two or more drivers are interacting. Finally,
a range of behaviors is defined for drivers and pedestrians: walking,
standing, looking, moving, etc.

An example of textual annotation is shown in \figref{Textual-annotation-example}.
The sequence of events recovered from this data is shown in \figref{timeline_of_crossing}.
\begin{figure*}
\noindent \begin{centering}
\includegraphics[scale=0.1]{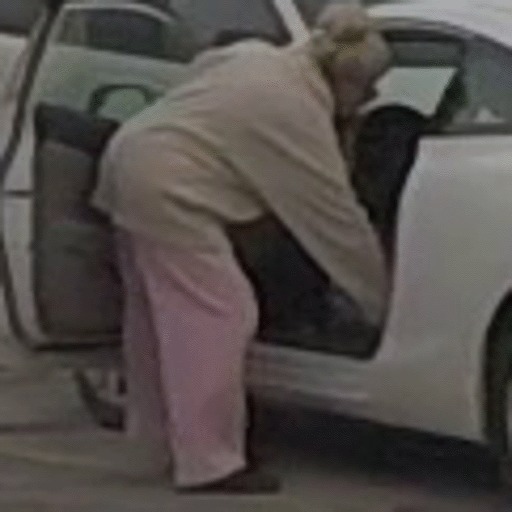}\includegraphics[scale=0.1]{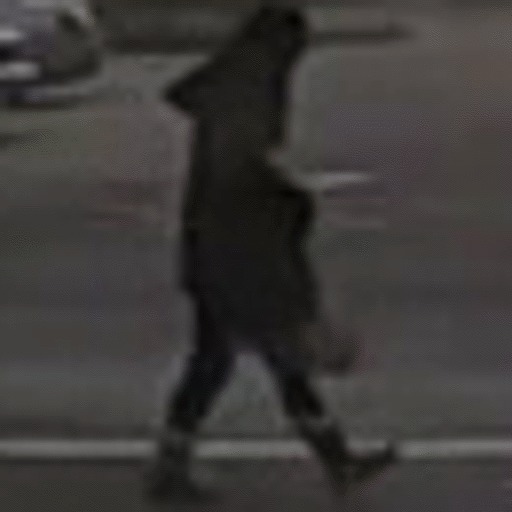}\includegraphics[scale=0.1]{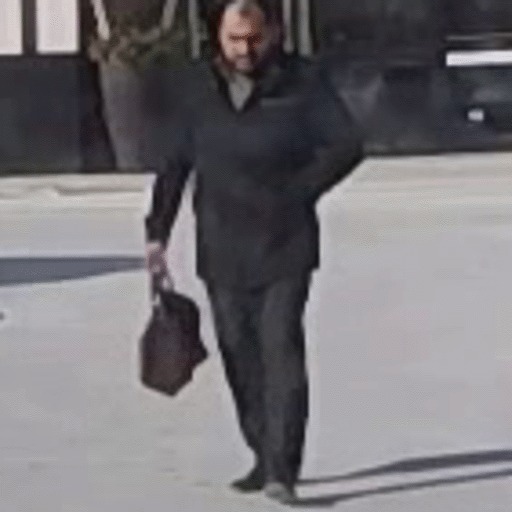}\includegraphics[scale=0.1]{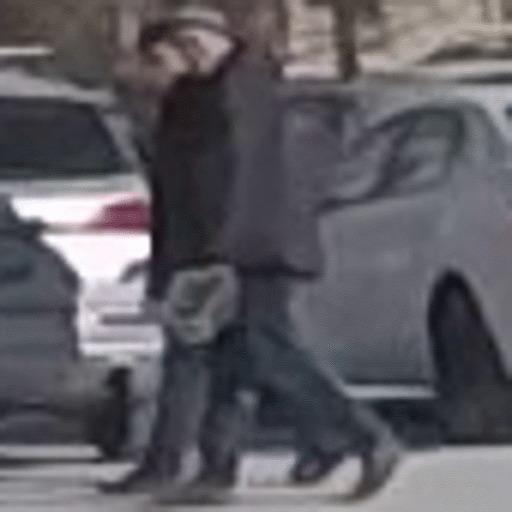}\includegraphics[scale=0.1]{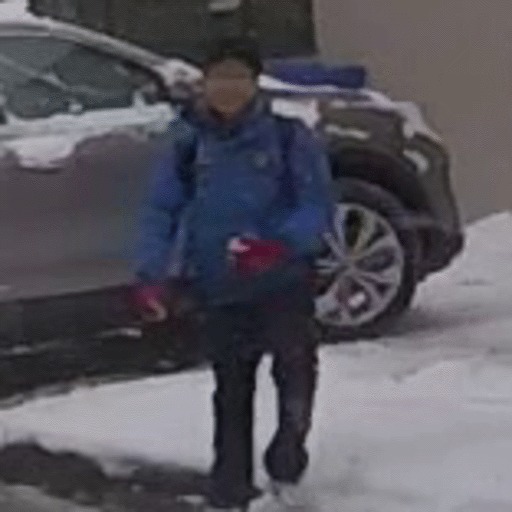}\includegraphics[scale=0.1]{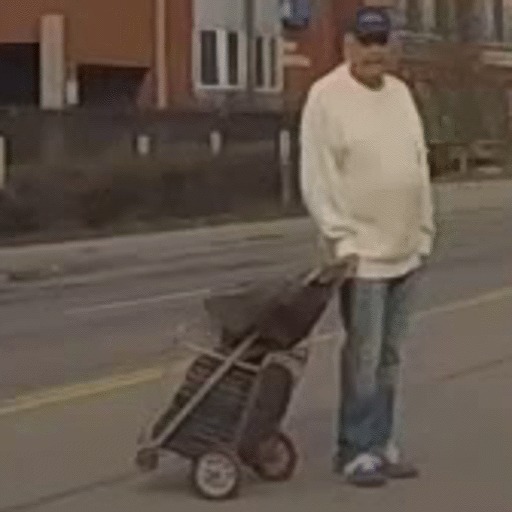}\includegraphics[scale=0.1]{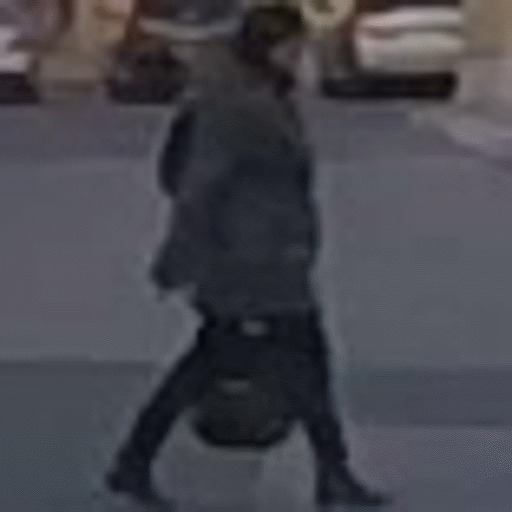}\includegraphics[scale=0.1]{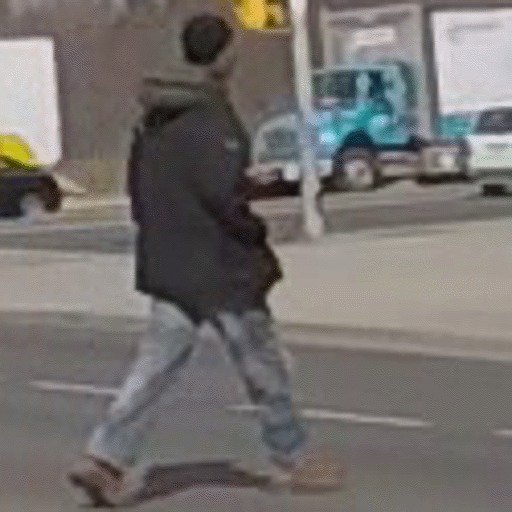}
\par\end{centering}
\noindent \begin{centering}
\includegraphics[scale=0.1]{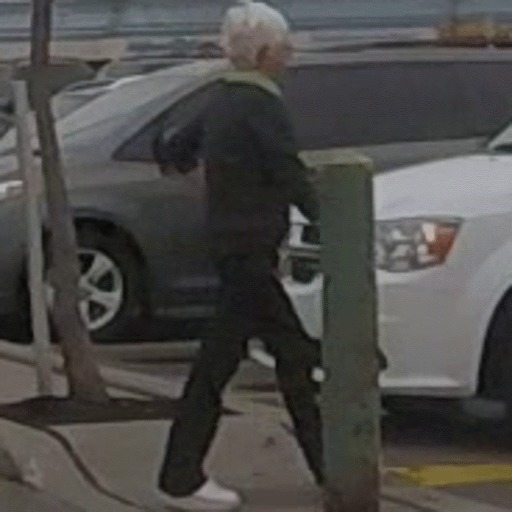}\includegraphics[scale=0.1]{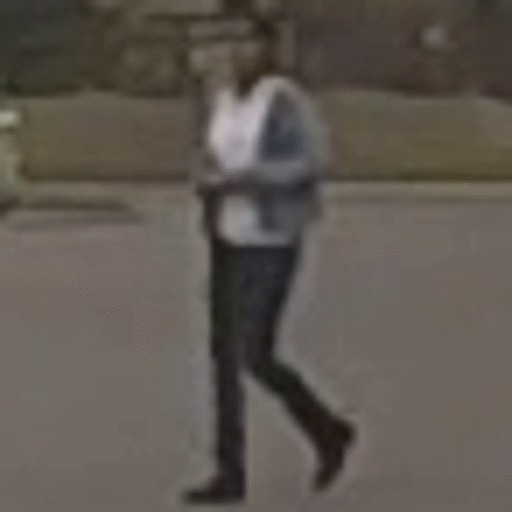}\includegraphics[scale=0.1]{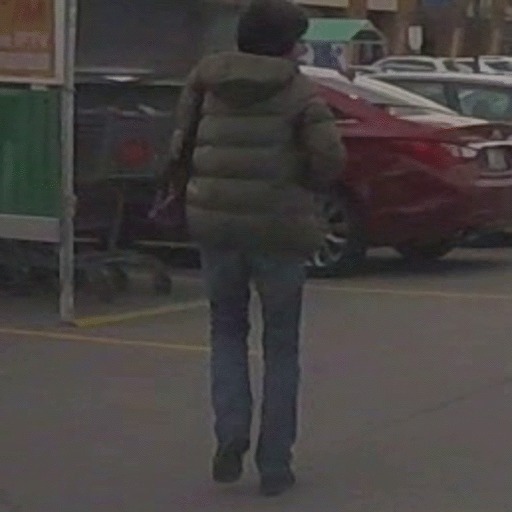}\includegraphics[scale=0.1]{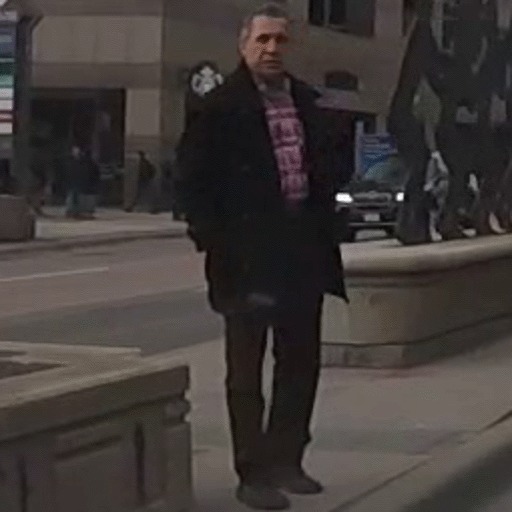}\includegraphics[scale=0.1]{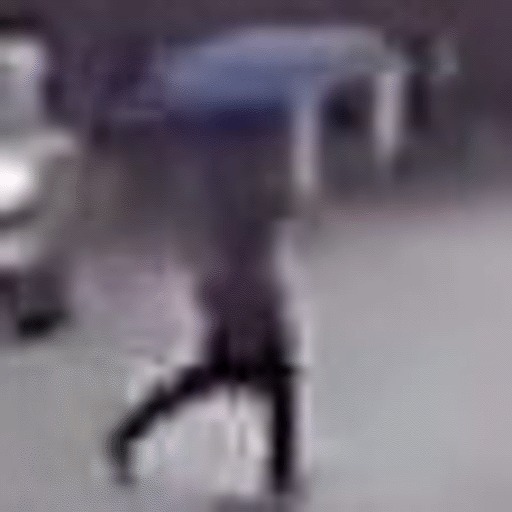}\includegraphics[scale=0.1]{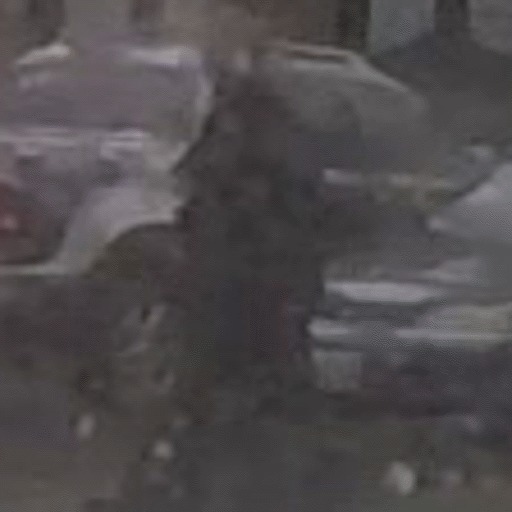}\includegraphics[scale=0.1]{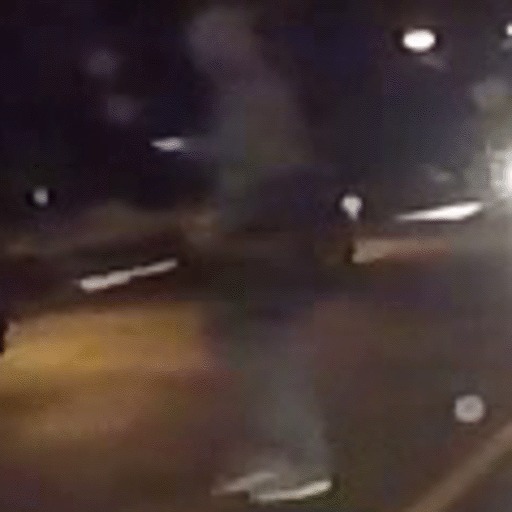}\includegraphics[scale=0.1]{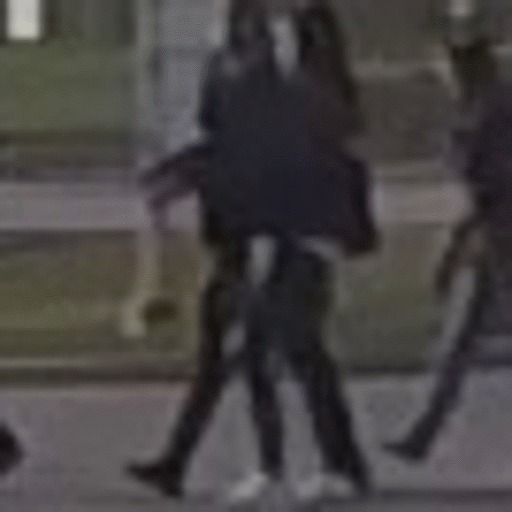}
\par\end{centering}
\noindent \begin{centering}
\includegraphics[scale=0.1]{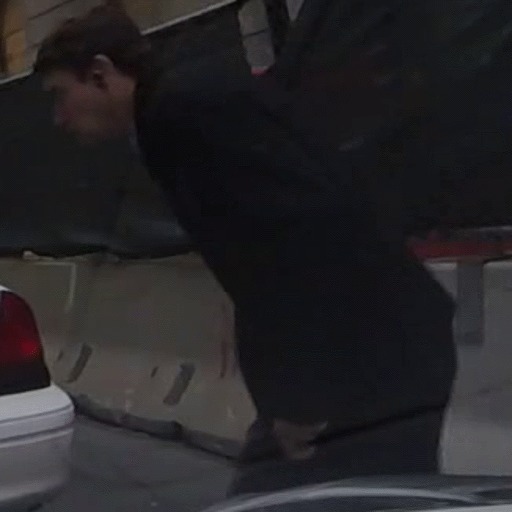}\includegraphics[scale=0.1]{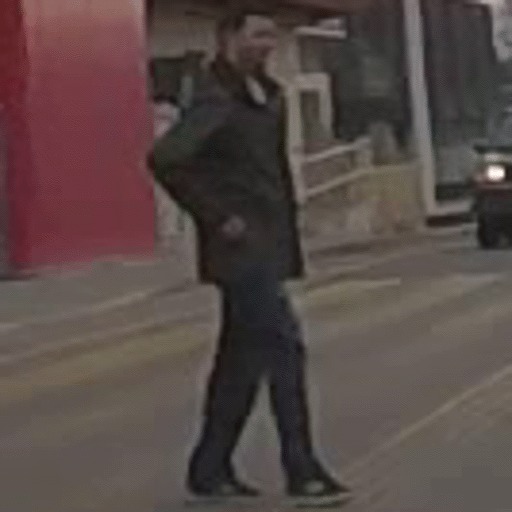}\includegraphics[scale=0.1]{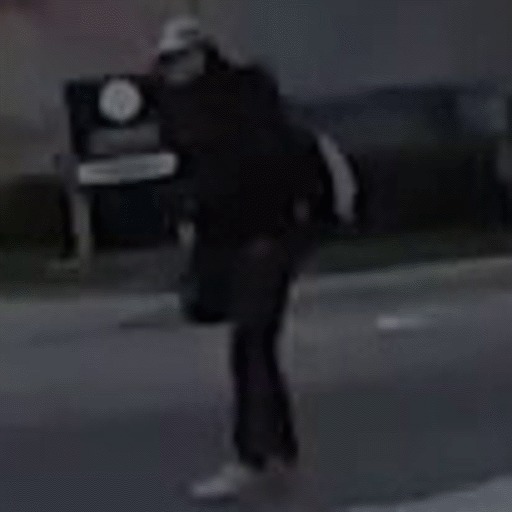}\includegraphics[scale=0.1]{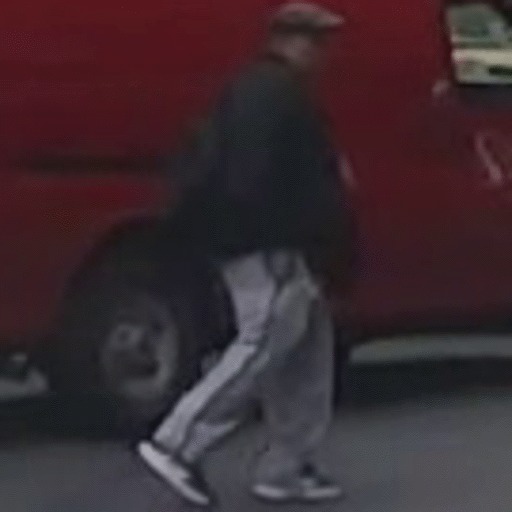}\includegraphics[scale=0.1]{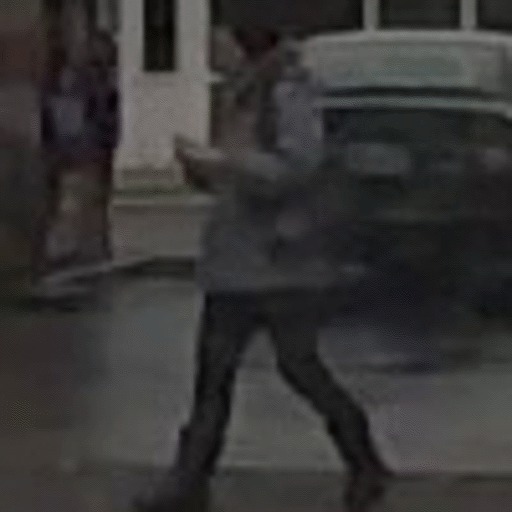}\includegraphics[scale=0.1]{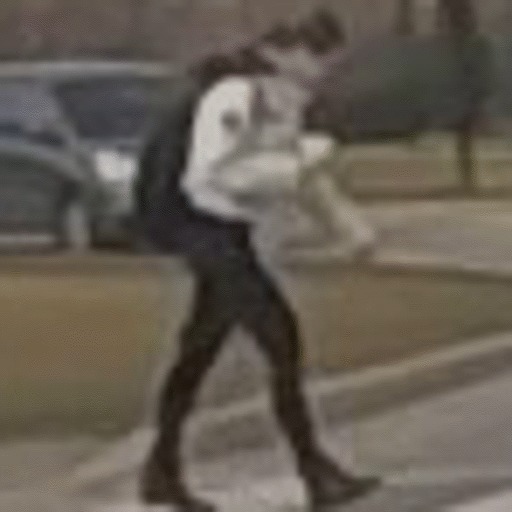}\includegraphics[scale=0.1]{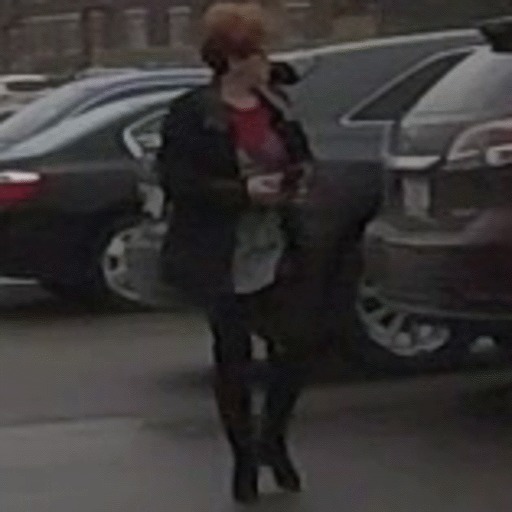}\includegraphics[scale=0.1]{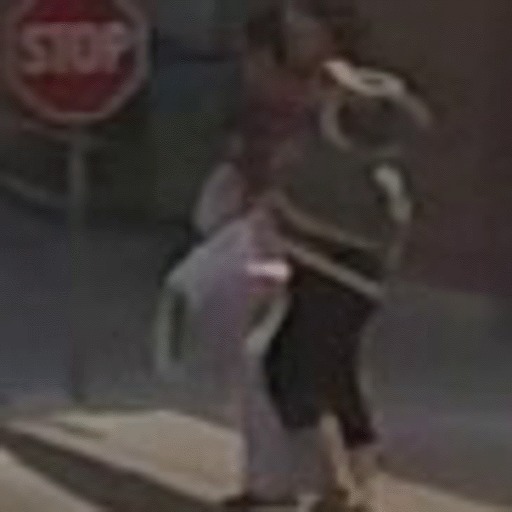}
\par\end{centering}
\noindent \begin{centering}
\includegraphics[scale=0.1]{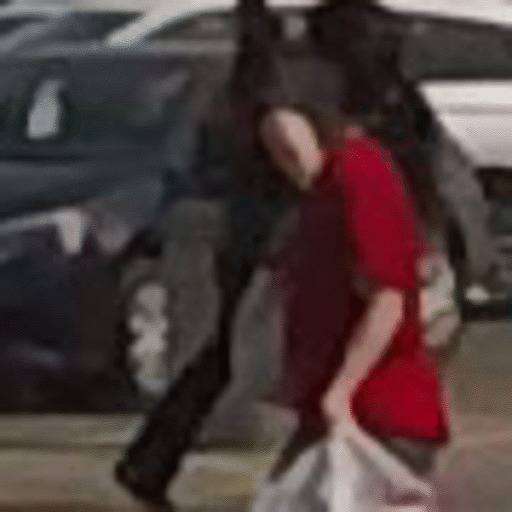}\includegraphics[scale=0.1]{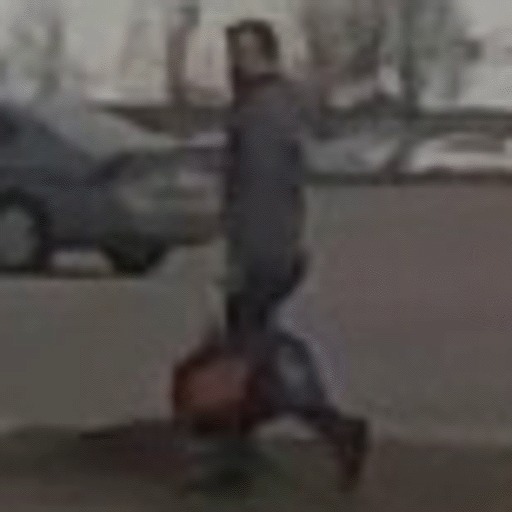}\includegraphics[scale=0.1]{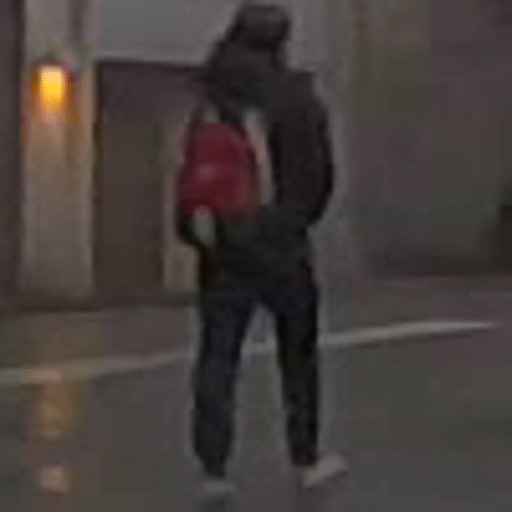}\includegraphics[scale=0.1]{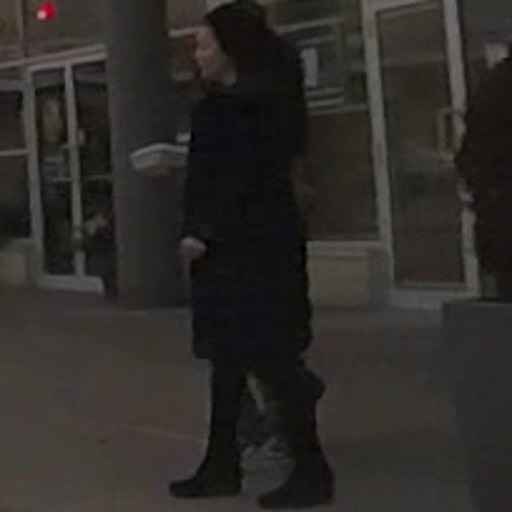}\includegraphics[scale=0.1]{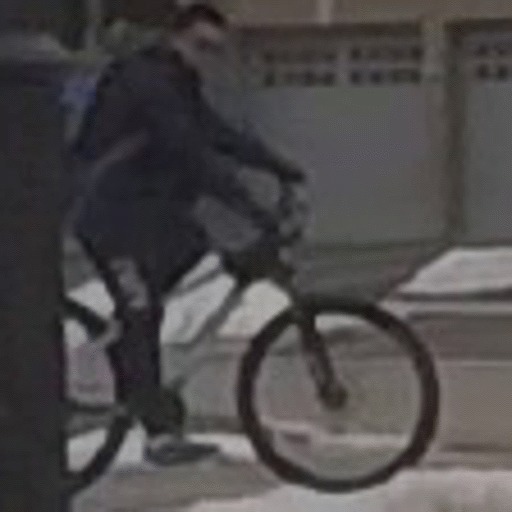}\includegraphics[scale=0.1]{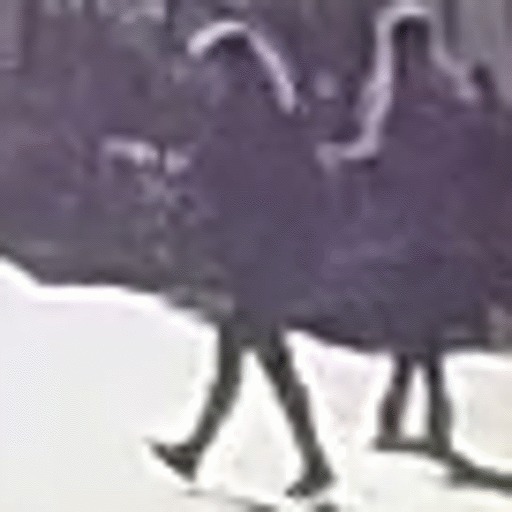}\includegraphics[scale=0.1]{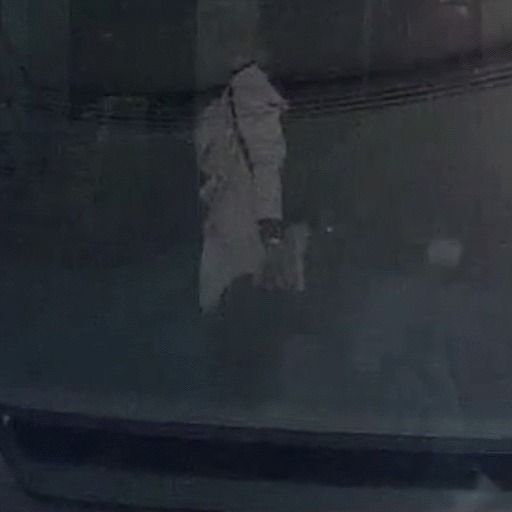}\includegraphics[scale=0.1]{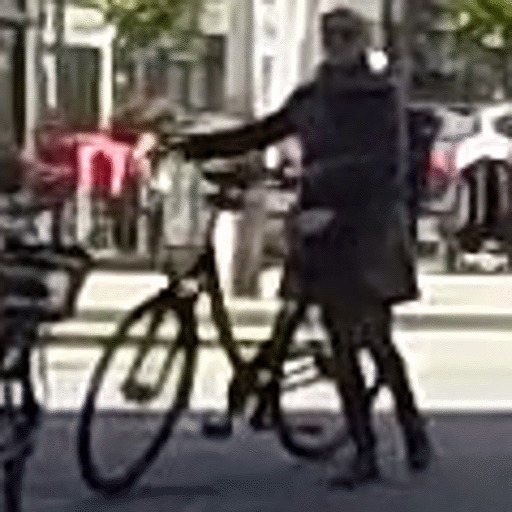}
\par\end{centering}
\noindent \begin{centering}
\includegraphics[scale=0.1]{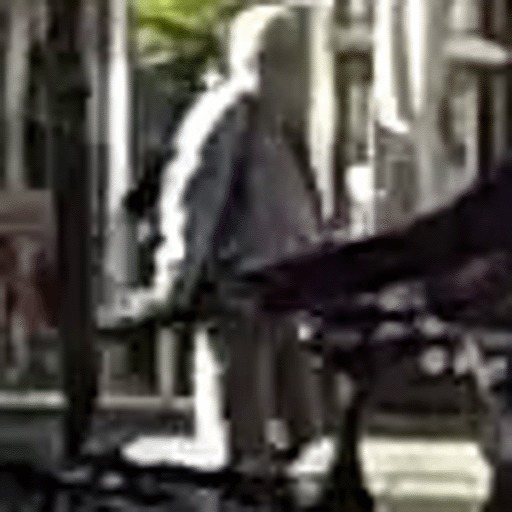}\includegraphics[scale=0.1]{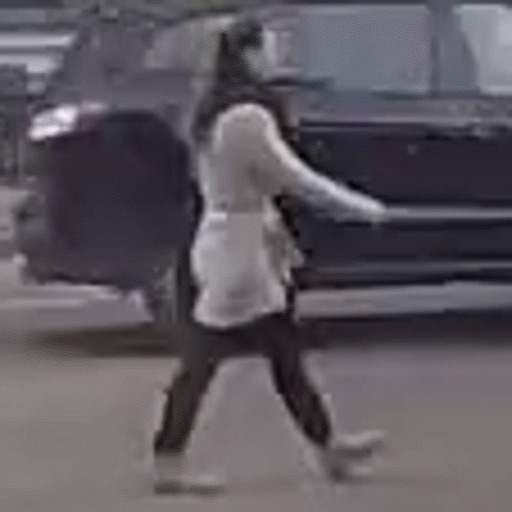}\includegraphics[scale=0.1]{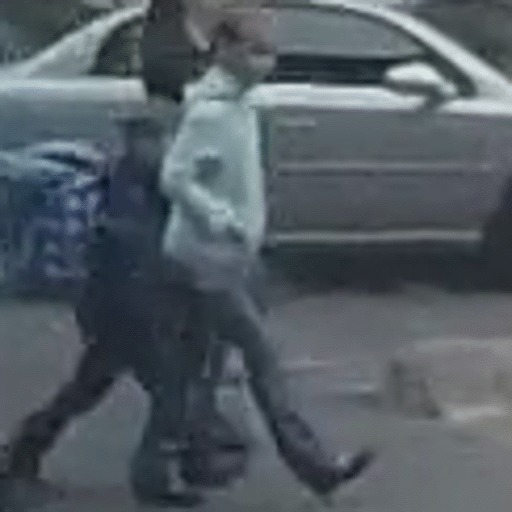}\includegraphics[scale=0.1]{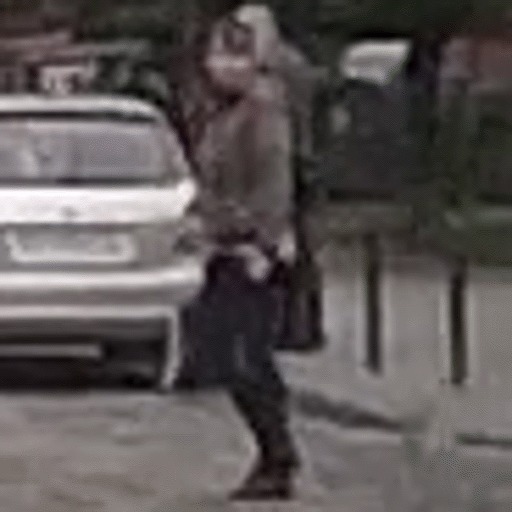}\includegraphics[scale=0.1]{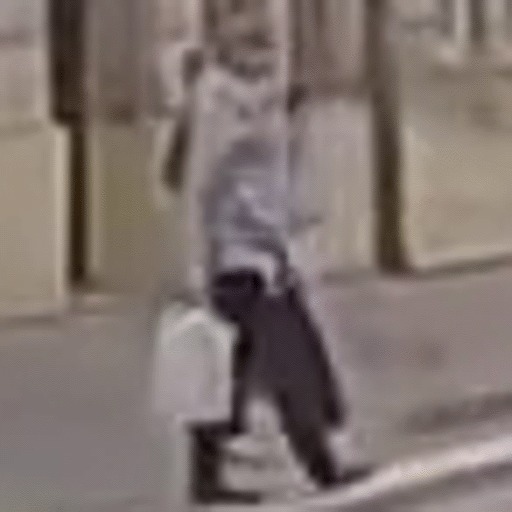}\includegraphics[scale=0.1]{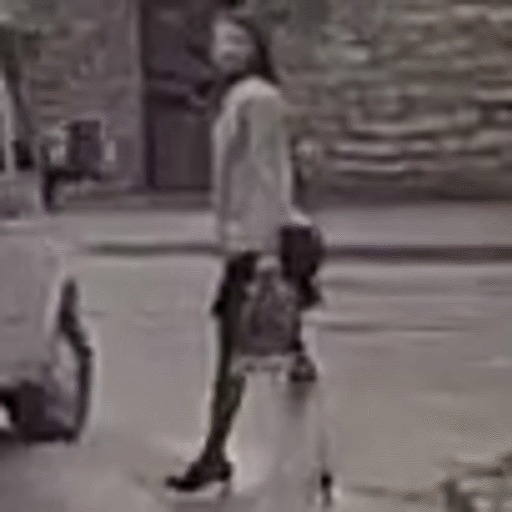}\includegraphics[scale=0.1]{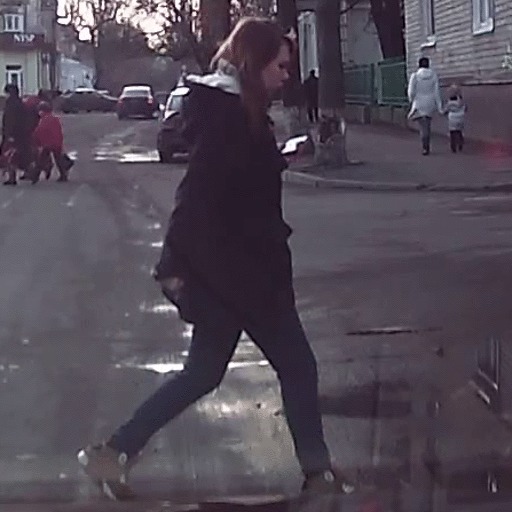}\includegraphics[scale=0.1]{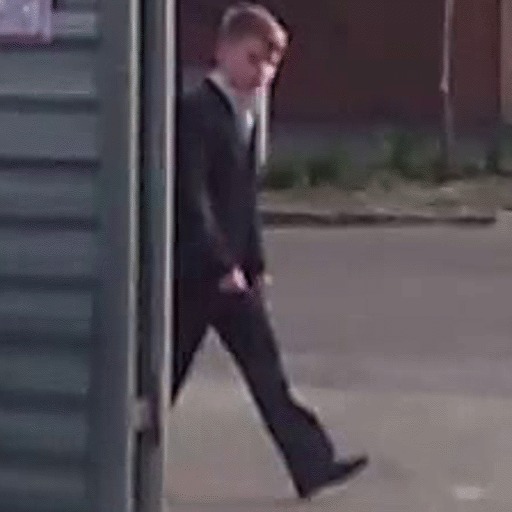}
\par\end{centering}
\noindent \begin{centering}
\includegraphics[scale=0.1]{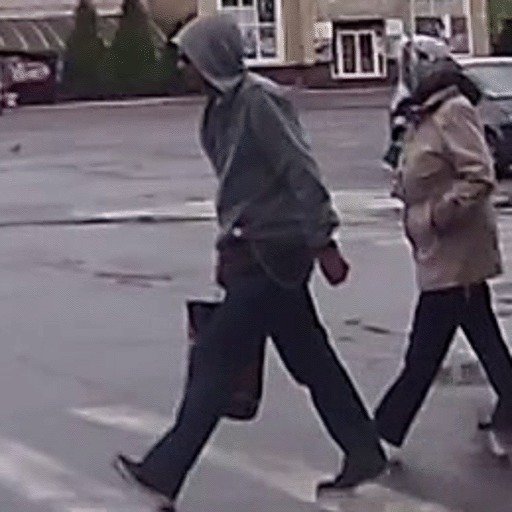}\includegraphics[scale=0.1]{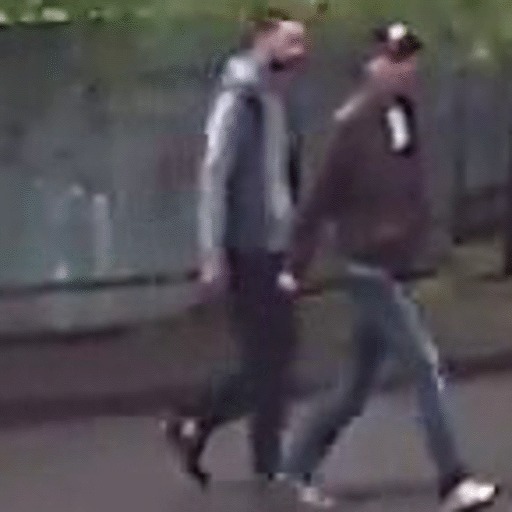}\includegraphics[scale=0.1]{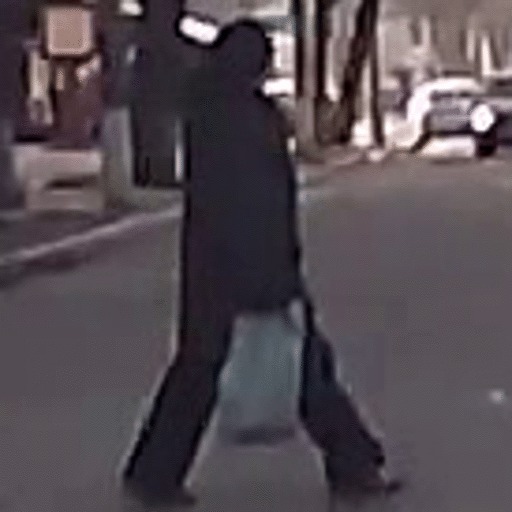}\includegraphics[scale=0.1]{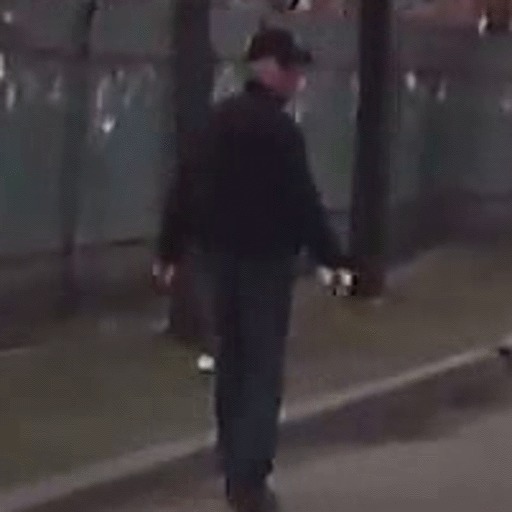}\includegraphics[scale=0.1]{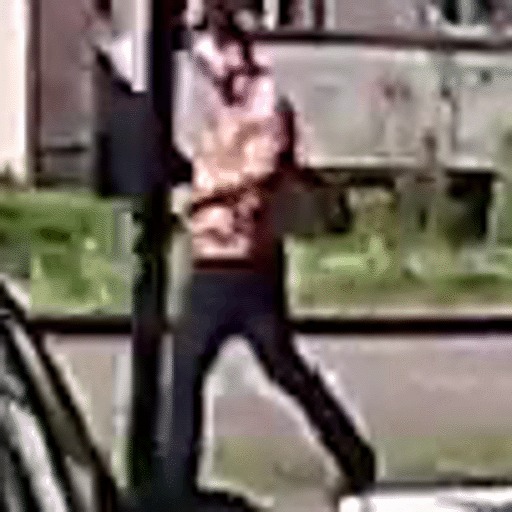}\includegraphics[scale=0.1]{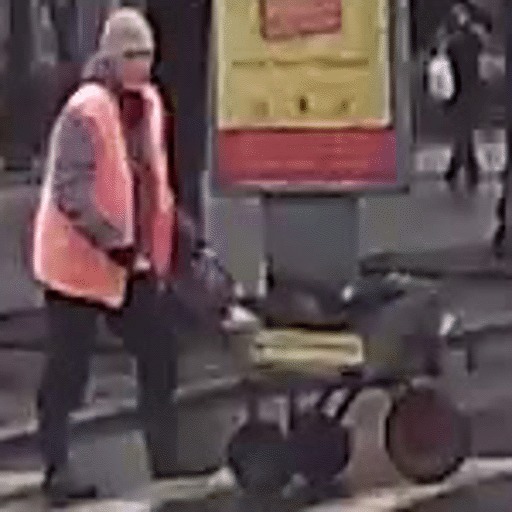}\includegraphics[scale=0.1]{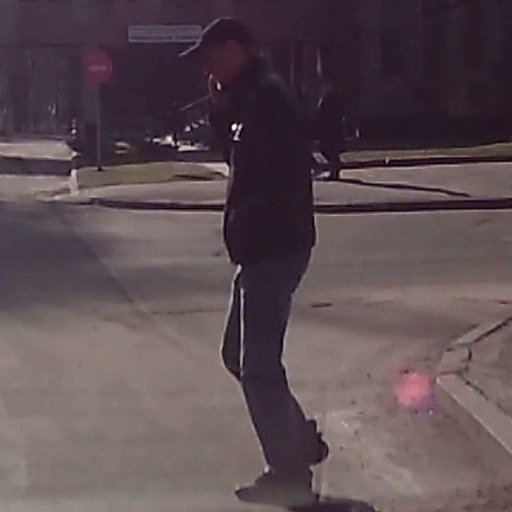}\includegraphics[scale=0.1]{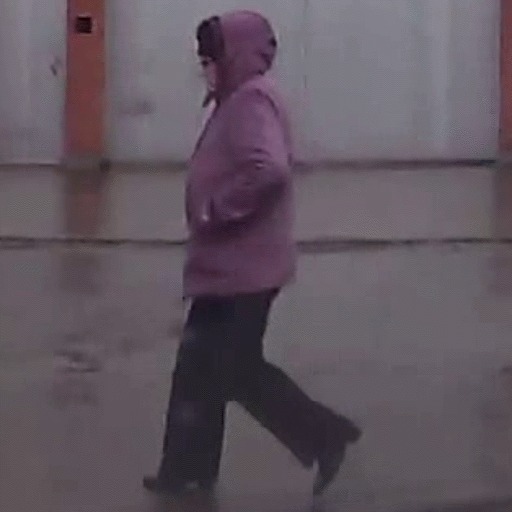}
\par\end{centering}
\noindent \begin{centering}
\includegraphics[scale=0.1]{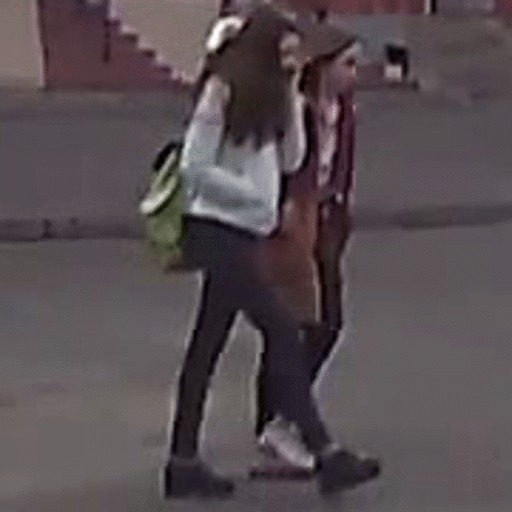}\includegraphics[scale=0.1]{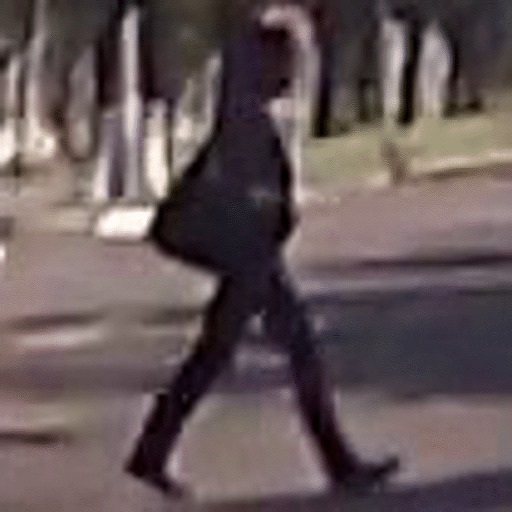}\includegraphics[scale=0.1]{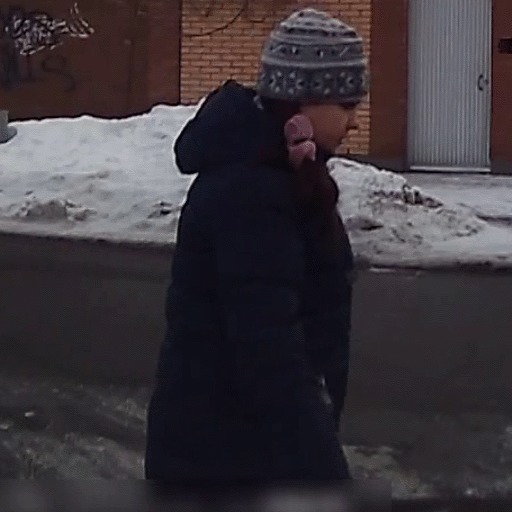}\includegraphics[scale=0.1]{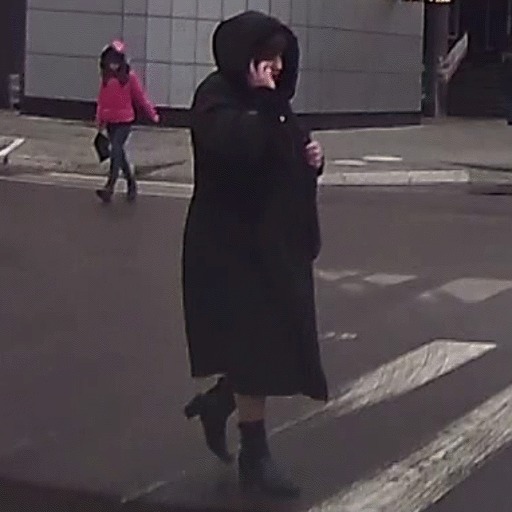}\includegraphics[scale=0.1]{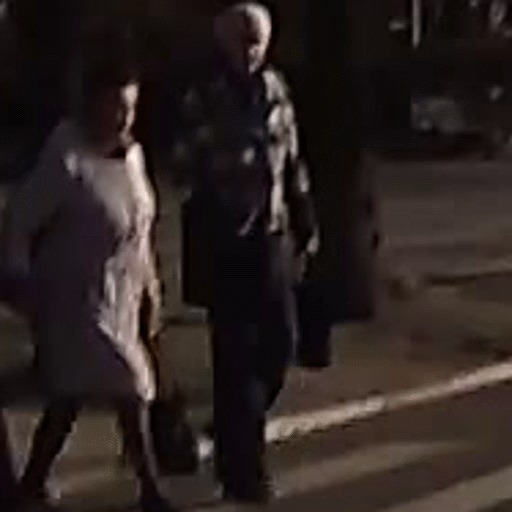}\includegraphics[scale=0.1]{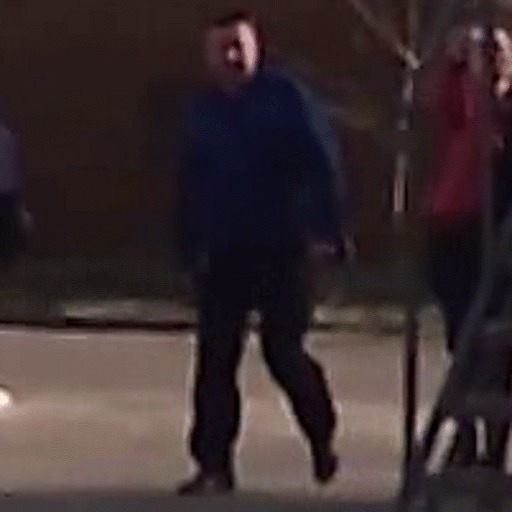}\includegraphics[scale=0.1]{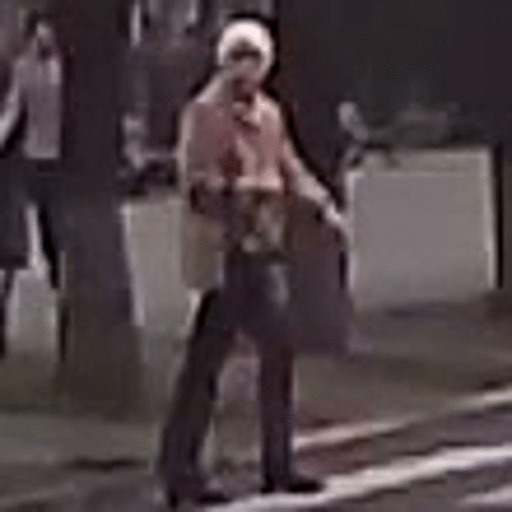}\includegraphics[scale=0.1]{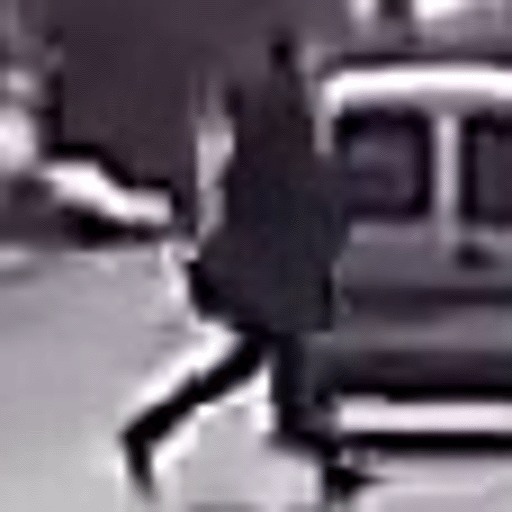}
\par\end{centering}
\noindent \begin{centering}
\includegraphics[scale=0.1]{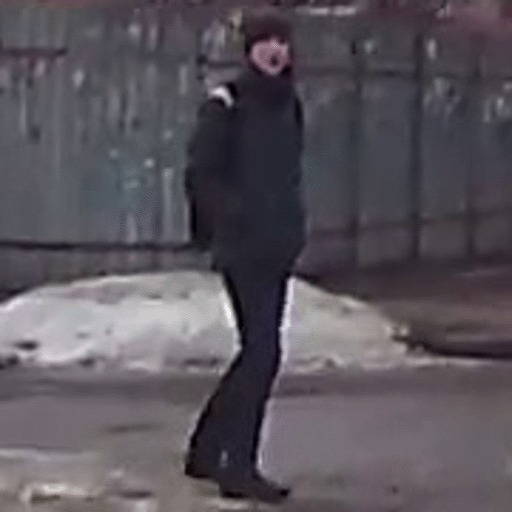}\includegraphics[scale=0.1]{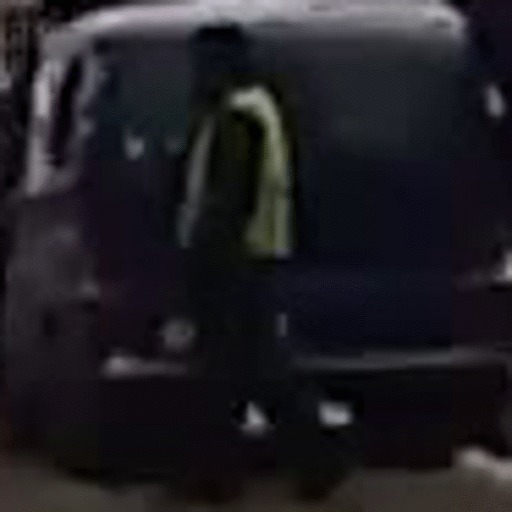}\includegraphics[scale=0.1]{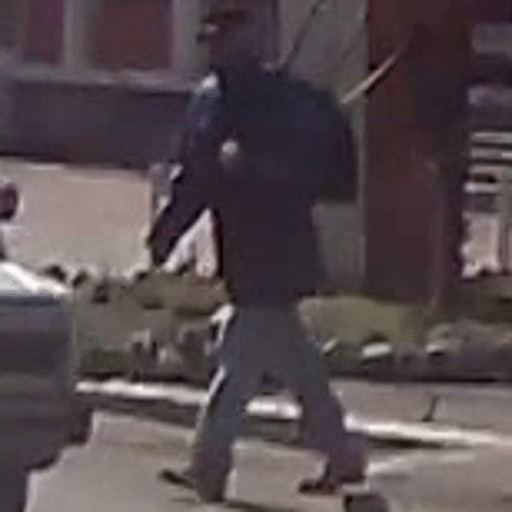}\includegraphics[scale=0.1]{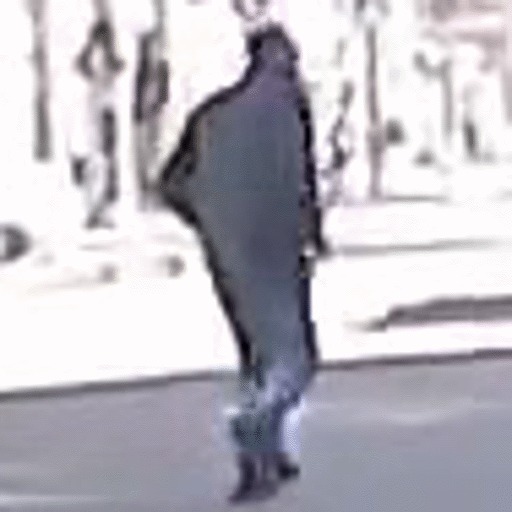}\includegraphics[scale=0.1]{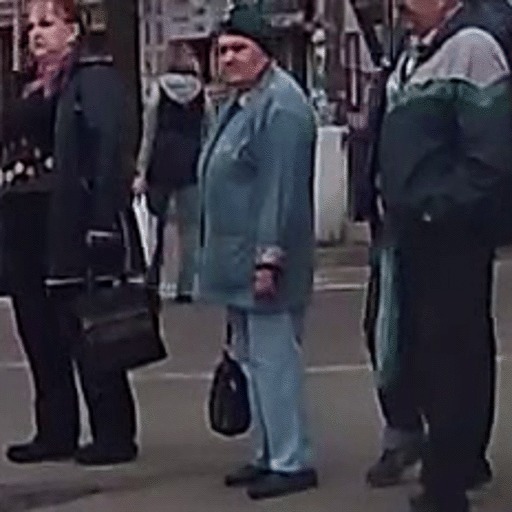}\includegraphics[scale=0.1]{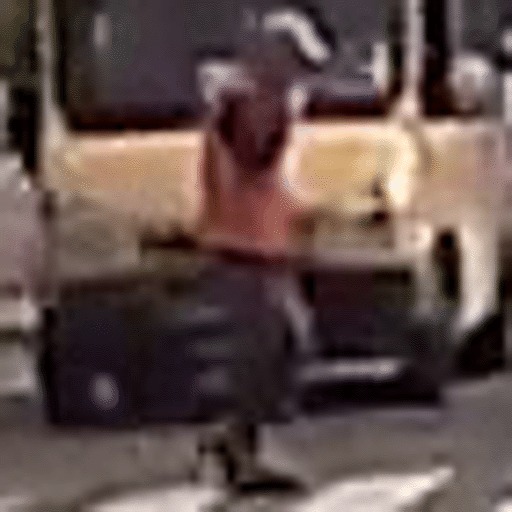}\includegraphics[scale=0.1]{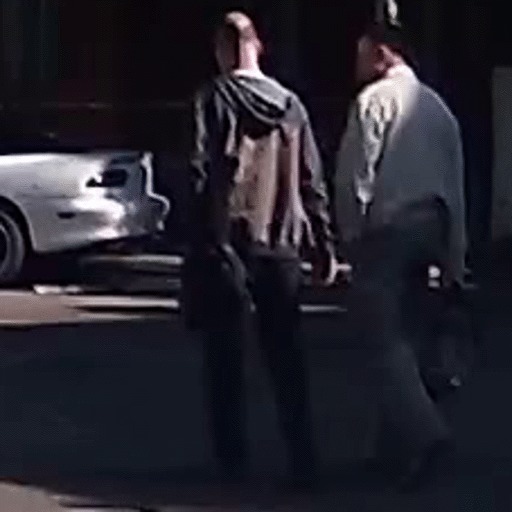}\includegraphics[scale=0.1]{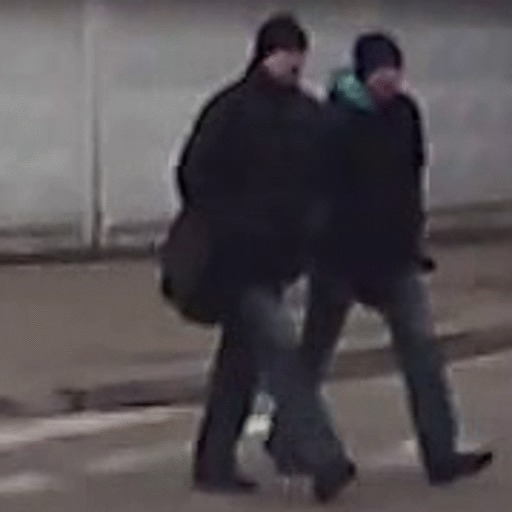}
\par\end{centering}
\noindent \begin{centering}
\includegraphics[scale=0.1]{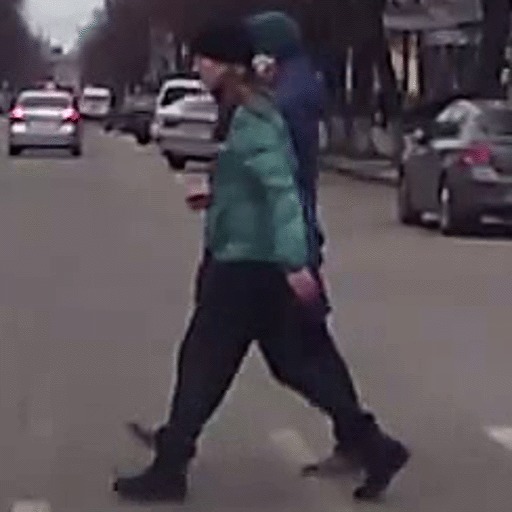}\includegraphics[scale=0.1]{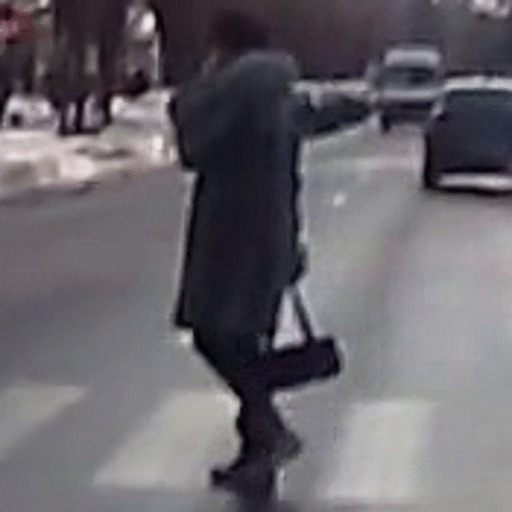}\includegraphics[scale=0.1]{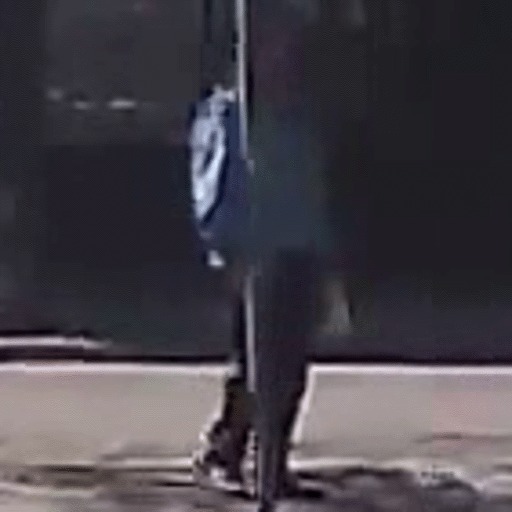}\includegraphics[scale=0.1]{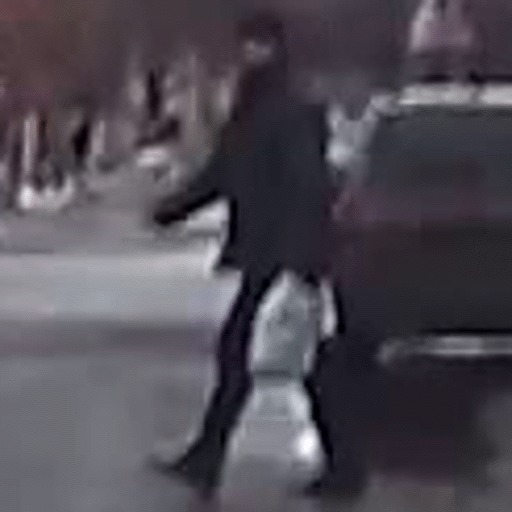}\includegraphics[scale=0.1]{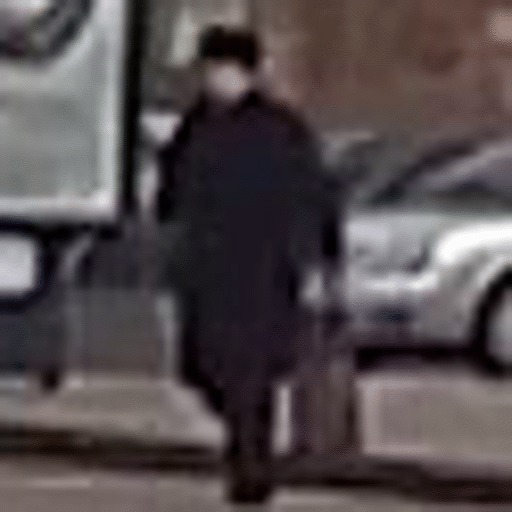}\includegraphics[scale=0.1]{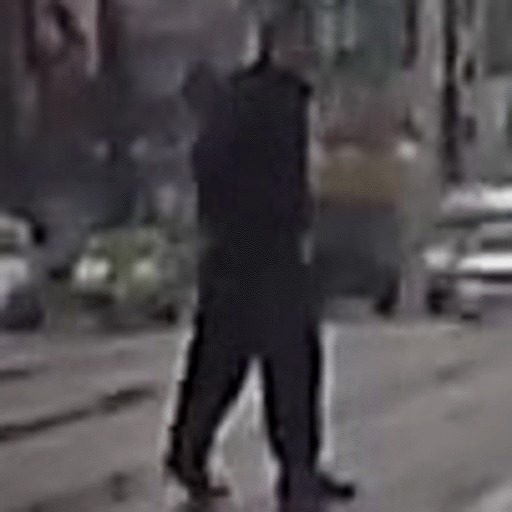}\includegraphics[scale=0.1]{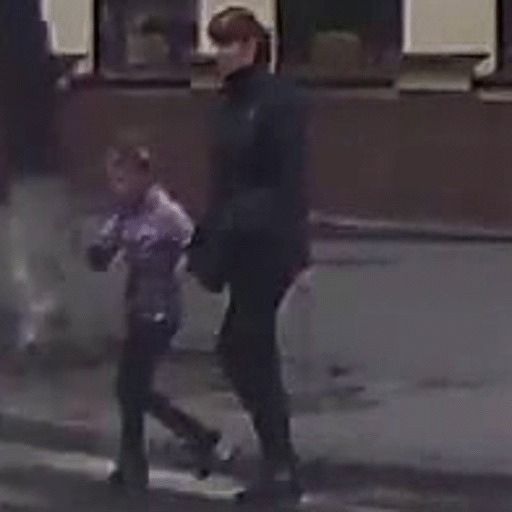}\includegraphics[scale=0.1]{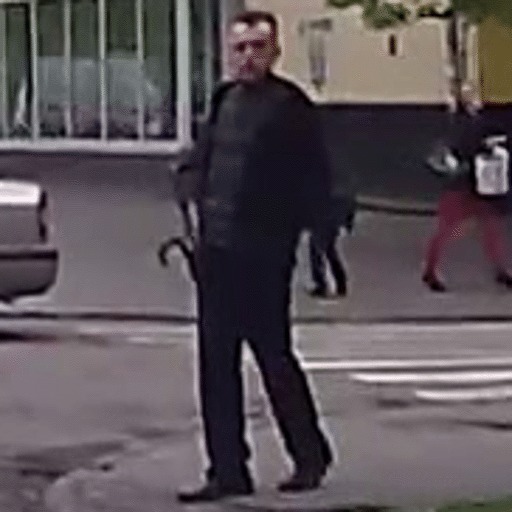}
\par\end{centering}
\caption{\label{fig:random-selection}A selection of images of pedestrians
from the dataset}

\end{figure*}

The dataset is available to download at \url{http://data.nvision2.eecs.yorku.ca/JAAD_dataset}.

\section{Conclusion}

In this paper we presented a new dataset for the purpose of studying
joint attention in the context of autonomous driving. Two types of
annotations accompanying each video clip in the dataset make it suitable
for pedestrian and car detection, as well as other areas of research,
which could benefit from studying joint attention and human non-verbal
communication, such as social robotics.

\section*{Acknowledgment}

We thank Mr. Viktor Kotseruba for assistance with processing videos
for this dataset.

\bibliographystyle{IEEEtran}
\bibliography{refs}

\begin{thebibliography}{10}
\providecommand{\url}[1]{#1}
\csname url@samestyle\endcsname
\providecommand{\newblock}{\relax}
\providecommand{\bibinfo}[2]{#2}
\providecommand{\BIBentrySTDinterwordspacing}{\spaceskip=0pt\relax}
\providecommand{\BIBentryALTinterwordstretchfactor}{4}
\providecommand{\BIBentryALTinterwordspacing}{\spaceskip=\fontdimen2\font plus
\BIBentryALTinterwordstretchfactor\fontdimen3\font minus
  \fontdimen4\font\relax}
\providecommand{\BIBforeignlanguage}[2]{{%
\expandafter\ifx\csname l@#1\endcsname\relax
\typeout{** WARNING: IEEEtran.bst: No hyphenation pattern has been}%
\typeout{** loaded for the language `#1'. Using the pattern for}%
\typeout{** the default language instead.}%
\else
\language=\csname l@#1\endcsname
\fi
#2}}
\providecommand{\BIBdecl}{\relax}
\BIBdecl

\bibitem{Litm2015}
\BIBentryALTinterwordspacing
T.~Litman. (2014, dec) {Autonomous Vehicle Implementation Predictions:
  Implications for Transport Planning}. Online. [Online]. Available:
  \url{http://www.vtpi.org/avip.pdf}
\BIBentrySTDinterwordspacing

\bibitem{Rol2014}
\BIBentryALTinterwordspacing
(2014, nov) {Think Act: Autonomous Driving}. Online. [Online]. Available:
  \url{https://new.rolandberger.com/wp-content/uploads/Roland{\_}Berger{\_}Autonomous-Driving1.pdf}
\BIBentrySTDinterwordspacing

\bibitem{SebastianThrun2006}
S.~Thrun, M.~Montemerlo, H.~Dahlkamp, D.~Stavens, A.~Aron, J.~Diebel, J.~G.
  {Philip Fong}, G.~H. {Morgan Halpenny}, K.~Lau, M.~P. {Celia Oakley},
  V.~Pratt, and P.~Stang, ``{Stanley: The Robot that Won the DARPA Grand
  Challenge},'' \emph{Journal of Field Robotics}, vol.~23, no.~9, pp. 661--692,
  2006.

\bibitem{Kalra2006}
\BIBentryALTinterwordspacing
N.~Kalra and S.~M. Paddock. (2006, apr) {Driving to Safety: How Many Miles of
  Driving Would It Take to Demonstrate Autonomous Vehicle Reliability}.
  [Online]. Available:
  \url{http://www.rand.org/pubs/research{\_}reports/RR1478.html}
\BIBentrySTDinterwordspacing

\bibitem{Knight2015}
\BIBentryALTinterwordspacing
W.~Knight. (2015, dec) {Can This Man Make AI More Human?} Online. [Online].
  Available:
  \url{https://www.technologyreview.com/s/544606/can-this-man-make-aimore-human/}
\BIBentrySTDinterwordspacing

\bibitem{Gomes2014}
\BIBentryALTinterwordspacing
L.~Gomes. (2014, jul) {Urban Jungle a Tough Challenge for Google's Autonomous
  Cars}. Online. [Online]. Available:
  \url{https://www.technologyreview.com/s/529466/urban-jungle-a-tough-challenge-for-googles-autonomous-cars/}
\BIBentrySTDinterwordspacing

\bibitem{Silberg2012}
\BIBentryALTinterwordspacing
G.~Silberg and R.~Wallace. (2012) {Self-driving cars: The next revolution}.
  Online. [Online]. Available:
  \url{https://www.kpmg.com/Ca/en/IssuesAndInsights/ArticlesPublications/Documents/self-driving-cars-next-revolution.pdf}
\BIBentrySTDinterwordspacing

\bibitem{Anthony2016}
\BIBentryALTinterwordspacing
S.~E. Anthony. (2016, mar) {The Trollable Self-Driving Car}. Online. [Online].
  Available:
  \url{http://www.slate.com/articles/technology/future{\_}tense/2016/03/}
\BIBentrySTDinterwordspacing

\bibitem{Richtel2016}
\BIBentryALTinterwordspacing
M.~Richtel and C.~Dougherty. (2016, sep) {Google's Driverless Cars Run Into
  Problem: Cars With Drivers}. Online. [Online]. Available:
  \url{http://www.nytimes.com/2015/09/02/technology/personaltech/google-says-its-not-the-driverless-cars-fault-its-other-drivers.html?{\_}r=1}
\BIBentrySTDinterwordspacing

\bibitem{Google2016}
\BIBentryALTinterwordspacing
(2016, feb) {Google Self-Driving Car Project Monthly Report}. Online. [Online].
  Available:
  \url{https://static.googleusercontent.com/selfdrivingcar/files/reports/report-0216.pdf}
\BIBentrySTDinterwordspacing

\bibitem{Knight2013}
\BIBentryALTinterwordspacing
W.~Knight. (2013, oct) {Driverless Cars Are Further Away Than You Think}.
  Online. [Online]. Available:
  \url{https://www.technologyreview.com/s/520431/driverless-cars-are-further-away-than-you-think/}
\BIBentrySTDinterwordspacing

\bibitem{Google2015}
\BIBentryALTinterwordspacing
(2015, dec) {Google Self-Driving Car Testing Report on Disengagements of
  Autonomous Mode}. Online. [Online]. Available:
  \url{https://static.googleusercontent.com/media/www.google.com/en//selfdrivingcar/files/reports/report-annual-15.pdf}
\BIBentrySTDinterwordspacing

\bibitem{V2V2014a}
\BIBentryALTinterwordspacing
W.~Knight. (2015) {Car-to-Car Communication}. Online. [Online]. Available:
  \url{https://www.technologyreview.com/s/534981/car-to-car-communication/}
\BIBentrySTDinterwordspacing

\bibitem{Ragland2007}
\BIBentryALTinterwordspacing
D.~R. Ragland and M.~F. Mitman, ``{Driver/Pedestrian Understanding and Behavior
  at Marked and Unmarked Crosswalks},'' Safe Transportation Research {\&}
  Education Center, Institute of Transportation Studies (UCB), UC Berkeley,
  Tech. Rep., 2007. [Online]. Available:
  \url{http://escholarship.org/uc/item/1h52s226}
\BIBentrySTDinterwordspacing

\bibitem{Honda2016a}
\BIBentryALTinterwordspacing
(2016, may) {Honda tech warns drivers of pedestrian presence}. Online.
  [Online]. Available:
  \url{http://www.cnet.com/roadshow/news/nikola-motor-company-the-ev-startup-with-the-worst-most-obvious-name-ever/}
\BIBentrySTDinterwordspacing

\bibitem{urmson2015pedestrian}
\BIBentryALTinterwordspacing
C.~P. Urmson, I.~J. Mahon, D.~A. Dolgov, and J.~Zhu, ``{Pedestrian
  notifications},'' 2015. [Online]. Available:
  \url{https://www.google.com/patents/US8954252}
\BIBentrySTDinterwordspacing

\bibitem{Stern2015}
\BIBentryALTinterwordspacing
G.~Stern. (2015, feb) {Robot Cars and Coordinated Chaos}. Online. [Online].
  Available:
  \url{http://www.wsj.com/articles/robot-cars-and-the-language-of-coordinated-chaos-1423540869}
\BIBentrySTDinterwordspacing

\bibitem{Jones2016}
\BIBentryALTinterwordspacing
R.~Jones. (2016, apr) {T3 Interview: Nissan's research chief talks autonomous
  vehicles and gunning it in his Nissan GT-R Black Edition}. Online. [Online].
  Available:
  \url{http://www.t3.com/features/t3-interview-dr-maarten-sierhuis-nissan-s-director-of-research-at-silicon-valley-talks-autonomous-vehicles-and-gunning-it-in-his-gt-r-black-edition}
\BIBentrySTDinterwordspacing

\bibitem{Ackerman2015}
\BIBentryALTinterwordspacing
E.~Ackerman and E.~Guizzo. (2015, sep) {Toyota Announces Major Push Into AI and
  Robotics, Wants Cars That Never Crash}. Online. [Online]. Available:
  \url{http://spectrum.ieee.org/automaton/robotics/artificial-intelligence/toyota-announces-major-push-into-ai-and-robotics}
\BIBentrySTDinterwordspacing

\bibitem{Akamatsu2003}
M.~Akamatsu, Y.~Sakaguchi, and M.~Okuwa, ``{Modeling of Driving Behavior When
  Approaching an Intersection Based on Measured Behavioral Data on An Actual
  Road},'' in \emph{Human factors and Ergonomics Society Annual Meeting
  Proceedings}, vol.~47, no.~16, 2003, pp. 1895--1899.

\bibitem{Fukagawa2013}
Y.~Fukagawa and K.~Yamada, ``{Estimating driver awareness of pedestrians from
  driving behavior based on a probabilistic model},'' in \emph{IEEE Intelligent
  Vehicles Symposium (IV)}, 2013.

\bibitem{Phan2014}
M.~T. Phan, I.~Thouvenin, V.~Fremont, and V.~Cherfaoui, ``{Estimating Driver
  Unawareness of Pedestrian Based On Visual Behaviors and Driving Behaviors},''
  in \emph{International Joint Conference on Computer Vision, Imaging and
  Computer Graphics Theory and Applications}, 2014.

\bibitem{Ren1026}
Z.~Ren, X.~Jiang, and W.~Wang, ``{Analysis of the Influence of Pedestrian's eye
  Contact on Driver's Comfort Boundary During the Crossing Conflict},''
  \emph{Green Intelligent Transportation System and Safety, Procedia
  Engineering}, 2016.

\bibitem{Gueguen2016}
N.~Gueguen, C.~Eyssartier, and S.~Meineri, ``{A pedestrian's smile and driver's
  behavior: When a smile increases careful driving},'' \emph{Journal of Safety
  Research}, vol.~56, pp. 83--88, 2016.

\bibitem{Kaplan2006a}
F.~Kaplan and V.~V. Hafner, ``{The Challenges of Joint Attention},''
  \emph{Interaction Studies}, vol.~7, no.~2, pp. 135--169, 2006.

\bibitem{Breazeal1999b}
\BIBentryALTinterwordspacing
C.~Breazeal and B.~Scassellati, ``{How to build robots that make friends and
  influence people},'' \emph{Intelligent Robots and Systems}, pp. 858--863,
  1999. [Online]. Available:
  \url{http://ieeexplore.ieee.org/xpls/abs{\_}all.jsp?arnumber=812787}
\BIBentrySTDinterwordspacing

\bibitem{Scassellati1999}
\BIBentryALTinterwordspacing
B.~Scassellati, ``{Imitation and Mechanisms of Joint Attention: A Developmental
  Structure for Building Social Skills on a Humanoid Robot},'' \emph{Lecture
  Notes in Computer Science}, vol. 1562, pp. 176--195, 1999. [Online].
  Available: \url{http://www.springerlink.com/content/wljp04e4h5b4lthh/}
\BIBentrySTDinterwordspacing

\bibitem{Shon2005}
A.~P. Shon, D.~B. Grimes, C.~L. Baker, M.~W. Hoffman, Z.~Shengli, and R.~P.~N.
  Rao, ``{Probabilistic gaze imitation and saliency learning in a robotic
  head},'' \emph{Proceedings - IEEE International Conference on Robotics and
  Automation}, vol. 2005, pp. 2865--2870, 2005.

\bibitem{Fasel2002}
I.~Fasel, G.~Deak, J.~Triesch, and J.~Movellan, ``{Combining embodied models
  and empirical research for understanding the development of shared
  attention},'' \emph{Proceedings 2nd International Conference on Development
  and Learning. ICDL 2002}, pp. 21--27, 2002.

\bibitem{Hafner2005}
V.~V. Hafner and F.~Kaplan, ``{Learning to interpret pointing gestures:
  Experiments with four-legged autonomous robots},'' \emph{Lecture Notes in
  Computer Science (including subseries Lecture Notes in Artificial
  Intelligence and Lecture Notes in Bioinformatics)}, vol. 3575 LNAI, pp.
  225--234, 2005.

\bibitem{Doniec2006}
\BIBentryALTinterwordspacing
M.~Doniec, G.~Sun, and B.~Scassellati, ``{Active Learning of Joint
  Attention},'' \emph{IEEE-RAS International Conference on Humanoid Robots},
  pp. 34--39, 2006. [Online]. Available:
  \url{http://ieeexplore.ieee.org/lpdocs/epic03/wrapper.htm?arnumber=4115577}
\BIBentrySTDinterwordspacing

\bibitem{Andry2001}
P.~Andry, P.~Gaussier, S.~Moga, J.~Banquet, and J.~Nadel, ``{Learning and
  Communication in Imitation: An Autnomous Robot Perspective},'' \emph{IEEE
  Transaction on Systems, Man and Cybernetics. Part A : Systems and Humans},
  vol.~31, no.~5, pp. 431--444, 2001.

\bibitem{Boucenna2011}
S.~Boucenna, P.~Gaussier, and L.~Hafemeister, ``{Development of joint attention
  and social referencing},'' \emph{2011 IEEE International Conference on
  Development and Learning, ICDL 2011}, pp. 1--6, 2011.

\bibitem{May2015}
A.~D. May, C.~Dondrup, and M.~Hanheide, ``{Show me your moves! Conveying
  navigation intention of a mobile robot to humans},'' \emph{Mobile Robots
  (ECMR), 2015 European Conference on}, pp. 1--6, 2015.

\bibitem{Ishiguro2002}
\BIBentryALTinterwordspacing
H.~Ishiguro, T.~Ono, M.~Imai, and T.~Maeda, ``{Robovie: an Interactive Humanoid
  Robot},'' \emph{IEEE International Conference}, vol.~2, no.~1, pp. 1848--
  1855, 2002. [Online]. Available:
  \url{http://onlinelibrary.wiley.com/doi/10.1002/cbdv.200490137/abstract$\backslash$nhttp://www.ingentaconnect.com/content/mcb/049/2001/00000028/00000006/art00006$\backslash$nhttp://ieeexplore.ieee.org/stamp/stamp.jsp?tp={\&}arnumber=1014810{\&}isnumber=21842}
\BIBentrySTDinterwordspacing

\bibitem{Ferreira2014}
J.~F. Ferreira and J.~Dias, ``{Attentional mechanisms for socially interactive
  robots - A survey},'' \emph{IEEE Transactions on Autonomous Mental
  Development}, vol.~6, no.~2, pp. 110--125, 2014.

\bibitem{Kelion2016}
\BIBentryALTinterwordspacing
L.~Kelion, ``{Tesla says autopilot involved in second car crash},'' jul 2016.
  [Online]. Available: \url{http://www.bbc.com/news/technology-36783345}
\BIBentrySTDinterwordspacing

\bibitem{Thielman2016}
\BIBentryALTinterwordspacing
S.~Thielman, ``{Fatal crash prompts federal investigation of Tesla self-driving
  cars},'' jul 2016. [Online]. Available:
  \url{https://www.theguardian.com/technology/2016/jul/13/tesla-autopilot-investigation-fatal-crash}
\BIBentrySTDinterwordspacing

\bibitem{Vincent2016}
\BIBentryALTinterwordspacing
J.~Vincent, ``{What counts as artificially intelligent? AI and deep learning,
  explained},'' \emph{The Verge}, feb 2016. [Online]. Available:
  \url{http://www.theverge.com/2016/2/29/11133682/deep-learning-ai-explained-machine-learning}
\BIBentrySTDinterwordspacing

\bibitem{Knight2015a}
\BIBentryALTinterwordspacing
W.~Knight, ``{Can This Man Make AI More Human},'' \emph{MIT Technology Review},
  dec 2015. [Online]. Available:
  \url{https://www.technologyreview.com/s/544606/can-this-man-make-aimore-human/}
\BIBentrySTDinterwordspacing

\bibitem{Goertzel2015}
B.~Goertzel, ``{Are there Deep Reasons Underlying the Pathologies of Today ' s
  Deep Learning Algorithms ?}'' \emph{Artificial General Intelligence Lecture
  Notes in Computer Science}, vol. 9205, pp. 70--79, 2015.

\bibitem{Szegedy2014}
\BIBentryALTinterwordspacing
C.~Szegedy, W.~Zaremba, and I.~Sutskever, ``{Intriguing properties of neural
  networks},'' in \emph{ICLR}, 2014, pp. 1--10. [Online]. Available:
  \url{http://arxiv.org/abs/1312.6199}
\BIBentrySTDinterwordspacing

\bibitem{Nguyen2015}
\BIBentryALTinterwordspacing
A.~Nguyen, J.~Yosinski, and J.~Clune, ``{Deep Neural Networks are Easily
  Fooled: High Confidence Predictions for Unrecognizable Images},'' in
  \emph{CVPR}, 2015. [Online]. Available: \url{http://arxiv.org/abs/1412.1897}
\BIBentrySTDinterwordspacing

\bibitem{KITTI2016b}
\BIBentryALTinterwordspacing
(2016, may) {The KITTI Vision Benchmark Suite}. Online. [Online]. Available:
  \url{http://www.cvlibs.net/datasets/kitti/}
\BIBentrySTDinterwordspacing

\bibitem{Dollar2012}
P.~Dollar, C.~Wojek, B.~Schiele, and P.~Perona, ``Pedestrian detection: An
  evaluation of the state of the art,'' \emph{PAMI}, vol.~34, 2012.

\bibitem{Fragkiadaki2012}
K.~Fragkiadaki, W.~Zhang, G.~Zhang, and J.~Shi, ``{Two-Granularity Tracking:
  Mediating Trajectory and Detection Graphs for Tracking under Occlusions},''
  in \emph{ECCV}, 2012.

\bibitem{Bao2012}
S.~Y. Bao, M.~Bagra, Y.-W. Chao, and S.~Savarese, ``{Semantic Structure from
  Motion with Points, Regions, and Objects},'' in \emph{CVPR}, 2012.

\bibitem{Sign2015a}
\BIBentryALTinterwordspacing
(2015) {The German Traffic Sign Detection Benchmark}. Online. [Online].
  Available:
  \url{http://benchmark.ini.rub.de/?section=gtsdb{\&}subsection=dataset}
\BIBentrySTDinterwordspacing

\bibitem{Klette2014}
\BIBentryALTinterwordspacing
R.~Klette. (2014) {The .enpeda.. Image Sequence Analysis Test Site (EISATS)}.
  Online. [Online]. Available:
  \url{http://ccv.wordpress.fos.auckland.ac.nz/eisats/{\#}1}
\BIBentrySTDinterwordspacing

\bibitem{Gavril2015}
\BIBentryALTinterwordspacing
D.~M. Gavrila. (2015) {Daimler Pedestrian Benchmark Datasets}. [Online].
  Available:
  \url{http://www.gavrila.net/Datasets/Daimler{\_}Pedestrian{\_}Benchmark{\_}D/daimler{\_}pedestrian{\_}benchmark{\_}d.html}
\BIBentrySTDinterwordspacing

\bibitem{Gavril2015a}
\BIBentryALTinterwordspacing
{UvA Person Tracking from Overlapping Cameras Dataset}. [Online]. Available:
  \url{http://www.gavrila.net/Datasets/Univ{\_}{\_}of{\_}Amsterdam{\_}Multi-Cam{\_}P/univ{\_}{\_}of{\_}amsterdam{\_}multi-cam{\_}p.html}
\BIBentrySTDinterwordspacing

\bibitem{100car}
\BIBentryALTinterwordspacing
{100-Car Naturalistic Driving Study}. [Online]. Available:
  \url{http://www.nhtsa.gov/Research/Human+Factors/Naturalistic+driving+studies:}
\BIBentrySTDinterwordspacing

\bibitem{110Car}
\BIBentryALTinterwordspacing
{Transportation Active Safety: 110-Car Naturalistic Driving Study}. [Online].
  Available:
  \url{http://www.engr.iupui.edu/{~}yidu/research.html{\#}Transportation}
\BIBentrySTDinterwordspacing

\bibitem{SHRP2}
\BIBentryALTinterwordspacing
{SHRP2 Naturalistic Driving Study}. [Online]. Available:
  \url{https://insight.shrp2nds.us/}
\BIBentrySTDinterwordspacing

\bibitem{VTTI}
\BIBentryALTinterwordspacing
{Virginia Tech Transportation Institute Data Warehouse}. [Online]. Available:
  \url{http://forums.vtti.vt.edu/index.php?/files/category/2-vtti-data-sets/}
\BIBentrySTDinterwordspacing

\bibitem{friard2016boris}
O.~Friard and M.~Gamba, ``Boris: a free, versatile open-source event-logging
  software for video/audio coding and live observations,'' \emph{Methods in
  Ecology and Evolution}, vol.~7, no.~11, pp. 1325--1330, 2016.

\end{thebibliography}

\end{document}